\documentclass[runningheads]{llncs}
\usepackage{array}

\usepackage{makeidx}
\usepackage{graphicx}
\usepackage{amsmath,amssymb} 
\usepackage{color}
\usepackage{multirow}
\usepackage{relsize}
\usepackage{comment}
\usepackage{pifont}

\begin{document}
	\pagestyle{headings}
	\mainmatter

    \title{Simultaneous Semantic Segmentation and
    \\
    Outlier Detection
    in Presence of Domain Shift} 
	\titlerunning{Semantic Segmentation and
    Outlier Detection
    in Presence of Domain Shift}
	\authorrunning{Bevandi{\' c} et al.}
\author{
Petra Bevandi{\' c}\thanks{This work has been partially supported by Croatian Science Foundation.} \and
Ivan Kre{\v s}o \and
Marin Or{\v s}i{\' c} \and
Sini{\v s}a {\v S}egvi{\' c}
}

\institute{
University of Zagreb, Faculty of Electrical Engineering and Computing,  Croatia
}

	\maketitle

\begin{abstract}
    Recent success on realistic
    road driving datasets
    has increased interest in exploring
    robust performance 
    in real-world applications.
    One of the major unsolved problems
    is to identify image content
    which can not be reliably recognized
    with a given inference engine.
    We therefore study approaches
    to recover a dense outlier map
    alongside the primary task
    with a single forward pass,
    by relying on shared convolutional features.
    We consider semantic segmentation
    as the primary task and
    perform extensive validation
    on WildDash val (inliers),
    LSUN val (outliers),
    and pasted objects from
    Pascal VOC 2007 (outliers).
    We achieve the best validation performance
    by training to discriminate inliers
    from pasted ImageNet-1k content,
    even though ImageNet-1k contains 
    many road-driving pixels,
    and, at least nominally, fails to account 
    for the full diversity of the visual world.
    The proposed two-head model performs
    comparably to the C-way multi-class model
    trained to predict 
    uniform distribution in outliers,
    while outperforming several
    other validated approaches.
    We evaluate our best two models 
    on the WildDash test dataset
    and set a new state of the art 
    on the WildDash benchmark.
\end{abstract}
\section{Introduction}

Early computer vision approaches focused
on producing decent performance 
on small datasets.
This often posed overwhelming difficulties, 
so researchers seldom quantified 
the prediction confidence. 
An important mi\-le\-stone
was reached when generalization 
was achieved
on realistic datasets
such as Pascal VOC \cite{everingham10ijcv},
CamVid \cite{brostow08eccv},
KITTI \cite{geiger13ijrr}, and
Cityscapes \cite{cordts15cvpr}.
These datasets assume 
closed-world evaluation 
\cite{scheirer13pami} 
in which the training and test subsets 
are sampled from the same distribution.
Such setup has been able to provide 
a fast feedback on novel approaches 
due to good alignment 
with the machine learning paradigm.
This further accelerated development 
and led us to the current state of research
where all these datasets are mostly solved, 
at least in the strongly supervised setup.

Recent datasets further raise the bar
by increasing the number of classes
and image diversity.
However, despite this increased complexity, 
the Vistas \cite{neuhold17iccv} dataset
is still an insufficient proxy 
for real-life operation
even in a very restricted scenario 
such as road driving.
New classes like bike racks 
and ground animals were added,
however many important classes from 
non-typical or worst-case images 
are still absent.
These classes include 
persons in non-standard poses,
crashed vehicles, rubble,
fallen trees etc.
Additionally, real-life images 
may be affected by 
particular image acquisition faults 
including hardware defects, 
covered lens etc.
This suggests that foreseeing every possible
situation may be an elusive goal, 
and that our algorithms
should be designed to recognize 
image regions which are foreign to
the training distribution.

\begin{figure}[htb]
  \includegraphics[width=\textwidth]{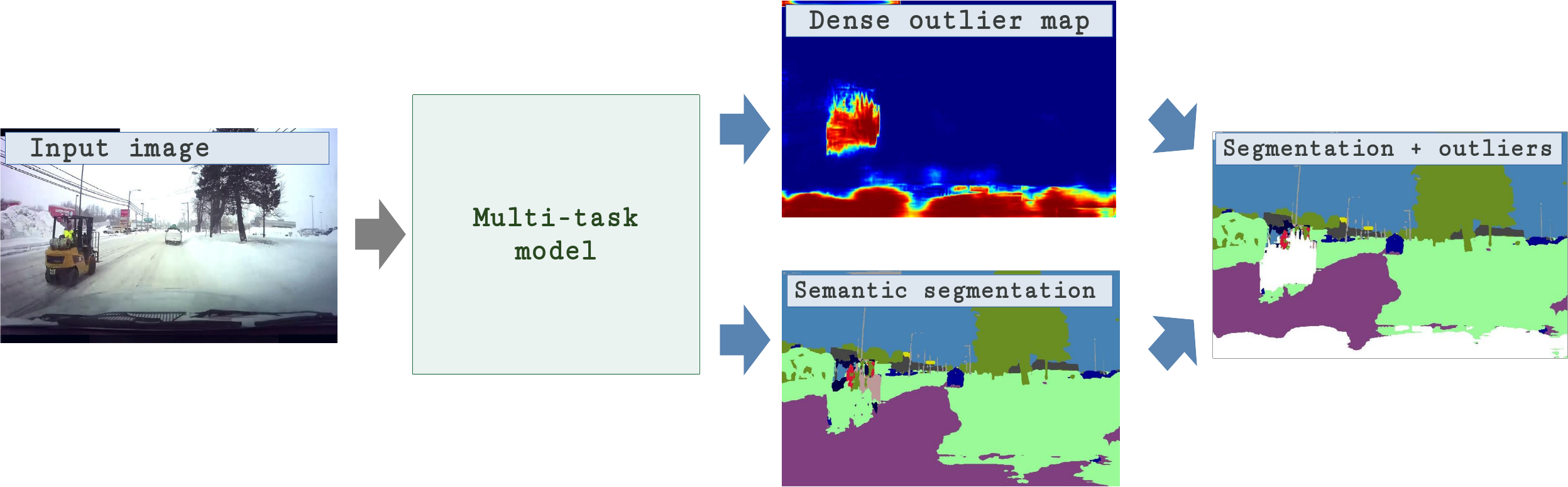}
  \caption{
    The proposed approach for simultaneous
    semantic segmentation and outlier detection.
    Our multi-task model predicts 
    i) a dense outlier map, and 
    ii) a semantic map with respect 
      to the 19 Cityscapes classes. 
    The two maps are merged to obtain
    the final outlier-aware 
    semantic predictions.
    Our model recognizes 
    outlier pixels (white) 
    on two objects
    which are foreign to Cityscapes:
    the ego-vehicle and the yellow forklift.
  }
  \label{fig:approach}
\end{figure}

The described deficiencies 
emphasize the need
for a more robust approach
to dataset design.
First, an ideal dataset should identify 
and target a set of explicit hazards
for the particular domain
\cite{zendel18eccv}.
Second (and more important),
an ideal dataset should endorse 
open-set recognition paradigm
\cite{scheirer13pami}
in order to promote 
detection of unforeseen hazards.
Consequently, the validation (val) 
and test subsets should contain
various degrees of domain shift
with respect to the training distribution.
This should include 
moderate domain shift factors
(e.g.\ adverse weather, exotic locations),
exceptional situations 
(e.g.\ accidents, poor visibility, defects)
and outright outliers 
(objects and entire images from other domains).
We argue that the WildDash dataset
\cite{zendel18eccv}
represents a step in the right direction,
although further development 
would be welcome, especially
in the direction of enlarging 
the negative part of the test dataset \cite{blum19arxiv}.
Models trained for open-set evaluation
can not be required 
to predict an exact visual class 
in outlier pixels.
Instead, it should suffice that 
the outliers are recognized,
as illustrated in Fig.\,\ref{fig:approach}
on an image from the WildDash dataset.

This paper addresses simultaneous 
semantic segmentation and 
open-set outlier detection.
We train our models on inlier images
from two road-driving datasets: 
Cityscapes \cite{cordts15cvpr} 
and Vistas \cite{neuhold17iccv}.
We consider several 
outlier detection approaches 
from the literature
\cite{hendrycks17iclr,bevandic2018,hendrycks19iclr}
and validate their performance
on WildDash val (inliers),
LSUN val \cite{yu2015} (outliers),
and pasted objects from
Pascal VOC 2007 (outliers).
Our main hypotheses are
i) that training with noisy negatives 
  from a very large and diverse dataset
  such as ImageNet-1k \cite{deng09cvpr}
  can improve outlier detection,
and ii) that discriminative outlier detection and semantic
segmentation can share features
without significant deterioration of either task.
We confirm both hypotheses
by re-training our best models 
on WildDash val, Vistas and ImageNet-1k,
and evaluating performance
on the WildDash benchmark.

\section{Related Work}

Previous approaches to 
outlier detection in image data
are very diverse.
These approaches are based on
analyzing prediction uncertainty,
evaluating generative models,
or exploiting a broad secondary dataset 
which contains both outliers and inliers.
Our approach is also related
to multi-task models 
and previous work which explores
the dataset quality and dataset bias.

\subsection{Estimating Uncertainty 
  (or Confidence) of the Predictions}
Prediction confidence can be expressed 
as the probability of the winning class 
or max-softmax for short \cite{hendrycks17iclr}.
This is useful in image-wide
prediction of outliers,
although max-softmax 
must be calibrated \cite{guo17icml}
before being interpreted as $P(\mathrm{inlier}|\mathbf{x})$.
The ODIN approach \cite{liang18iclr}
improves on \cite{hendrycks17iclr} 
by pre-processing input images 
with a well-tempered perturbation 
aimed at increasing 
the max-softmax activation.
These approaches are handy since they 
require no additional training.

Some approaches model the uncertainty
with a separate head which learns either
prediction uncertainty 
\cite{kendall17nips,lakshminarayanan17nips} 
or confidence \cite{devries18arxiv}.
Such training is able to recognize examples
which are hard to classify 
due to insufficient or inconsistent labels,
but is unable to deal with real outliers.

A principled information-theoretic approach 
expresses the prediction uncertainty as  
mutual information between 
the posterior parameter distribution
and the particular prediction \cite{smith18uai}.
In practice, the required expectations
are estimated with Monte Carlo (MC) dropout
\cite{kendall17nips}.
Better results 
have been achieved
with explicit ensembles
of independently trained models
\cite{lakshminarayanan17nips}.
However, both approaches 
require many forward passes
and thus preclude real-time operation.

Prediction uncertainty 
can also be expressed
by evaluating per-class 
generative models 
of latent features \cite{Lee2018ASU}.
However, this idea is not easily adaptable 
for dense prediction in 
which latent features 
typically correspond to many classes
due to subsampling and dense labelling.
Another approach would be 
to fit a generative model 
to the training dataset and 
to evaluate the likelihood of a given sample.
Unfortunately, this is very hard
to achieve with image data
\cite{nalisnick19iclr}. 

\subsection{Training with Negative Data}

Our approach is most related 
to three recent approaches 
which train outlier detection 
by exploiting a diverse 
negative dataset \cite{torralba2011}.
The approach called outlier exposure
(OE) \cite{hendrycks19iclr} 
processes the negative data
by optimizing cross entropy 
between the predictions 
and the uniform distribution.
Outlier detection has also been formulated
as binary classification \cite{bevandic2018}
trained to differentiate inliers 
from the negative dataset.
A related approach \cite{Vyas2018OutofDistributionDU} 
partitions the training data into K folds
and trains an ensemble of K
leave-one-fold-out classifiers.
However, this requires K forward passes.
while data partitioning 
may not be straight-forward.

Negative training samples
can also be produced 
by a GAN generator \cite{goodfellow14nips}.
Unfortunately, existing works 
\cite{lee18iclr,sabokrou2018adversarially}
have been designed for 
image-wide prediction in small images.
Their adaptation to dense prediction
on Cityscapes resolution
would not be straight-forward
\cite{brock19iclr}.

Soundness of training with negative data
has been challenged by \cite{shafaei2018}
who report under-average results 
for this approach.
However, their experiments average results 
over all negative datasets (including MNIST),
while we advocate for a very diverse  
negative dataset such as ImageNet.

        
\subsection{Multi-task Training 
  and Dataset Design}

Multi-task models attach 
several prediction heads
to shared features \cite{Caruana1997}.
Each prediction head has a distinct loss.
The total loss is usually expressed 
as a weighted sum \cite{ngiam2011}
and optimized in an end-to-end fashion.
Feature sharing brings important advantages 
such as cross-task enrichment
of training data \cite{bengio13pami}
and faster evaluation.
Examples of successful multi-task models
include combining
depth, surface normals and 
semantic segmentation 
\cite{Eigen2015PredictingDS},
as well as combining
classification, bounding box prediction 
and per-class instance-level segmentation
\cite{mask}.
A map of task compatibility 
with respect to knowledge transfer
\cite{zamir18cvpr}
suggests that many tasks are
suitable for multi-task training.

Dataset quality is  
as a very important issue 
in computer vision research.
Diverse negative datasets 
have been used to reduce false positives
in several computer vision tasks
for a very long time \cite{torralba2011}.
A methodology for analyzing 
the quality of road-driving datasets
has been proposed in \cite{zendel17ijcv}.
The WildDash dataset \cite{zendel18eccv} 
proposes a very diverse validation dataset
and the first semantic segmentation benchmark
with open-set evaluation \cite{scheirer13pami}.

\section{Simultaneous Segmentation and Outlier Detection}
\label{ss:model}

Our method combines two distinct tasks:
outlier detection and semantic segmentation,
as shown in Fig.\,\ref{fig:approach}. 
We prefer to rely on shared features 
in order to promote fast inference
and synergy between tasks \cite{bengio13pami}.
We assume that a large, diverse and noisy
negative dataset is available
for training purposes
\cite{hendrycks19iclr,bevandic2018}.

\subsection{Dense Feature Extractor}

\begin{figure}[htb]
  \centering
  \includegraphics[width=0.9\textwidth]{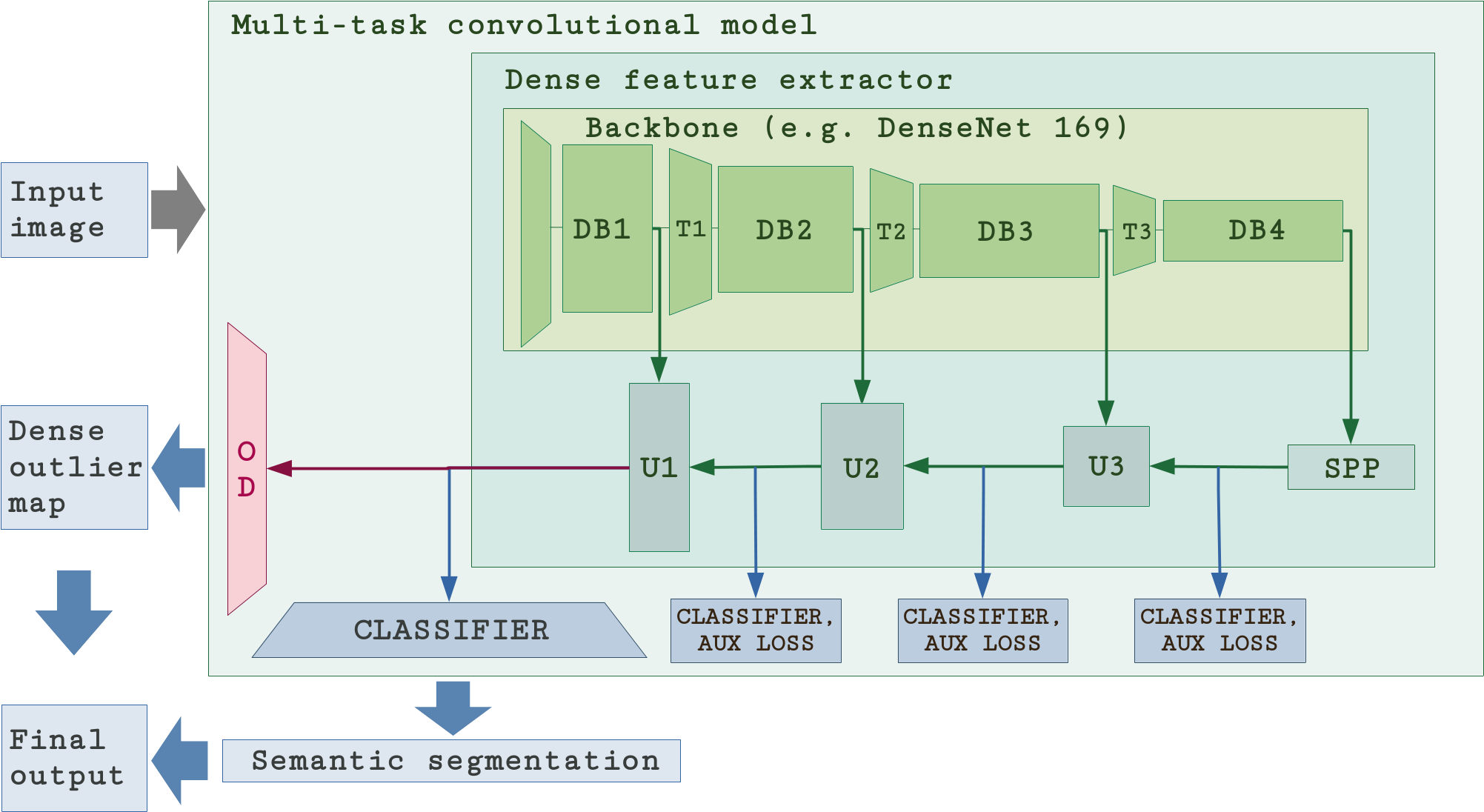}
  \caption{The proposed two-head model:
  the classification head recovers semantic segmentation
  while the outlier detection head 
  identifies pixels where 
  semantic segmentation may be wrong. 
  The output is produced by combining
  these two dense prediction maps. 
  }
  \label{fig:model}
  
\end{figure}

Our models are based on  
a dense feature extractor
with lateral connections
\cite{kreso17cvrsuad}.
The processing starts with a 
DenseNet \cite{huang17cvpr} or 
ResNet \cite{He2016DeepRL} backbone, 
proceeds with spatial pyramid pooling (SPP) 
\cite{he14eccv,zhao17cvpr}
and concludes with ladder-style upsampling
\cite{kreso17cvrsuad,lintsungyi17cvpr}.
The upsampling path consists of 
three upsampling blocks (U1-U3)
which blend low resolution features
from the previous upsampling stage
with high resolution features from the backbone.
We speed-up and regularize the learning
with three auxiliary classification losses
(cf.~Fig.\,\ref{fig:model}). 
These losses have soft targets corresponding
to ground truth distribution 
across the corresponding window
at full resolution \cite{kreso19arxiv}.

\subsection{Dense Outlier Detection}

There are four distinct approaches to formulate
simultaneous semantic segmentation and
dense outlier detection 
over shared features.
The C-way multi-class approach
attaches the standard classification head 
to the dense feature extractor
(C denotes the number of inlier classes).
The inlier probability 
is formulated as max-softmax.
If a negative set is available,
this approach can be trained 
to emit low max-softmax in outliers
by supplying a modulated 
cross entropy loss term
towards uniform distribution
\cite{lee18iclr,hendrycks19iclr}.
The modulation factor 
$\lambda_\textrm{KL}$
is a hyper-parameter. 
Unfortunately, training on outliers 
may compromise classification accuracy 
and generate false positive outliers
at semantic borders.

The C-way multi-label approach 
has C sigmoid heads.
The final prediction is the class 
with maximum probability max-$\sigma$,
whereas the inlier probability 
is formulated as max-$\sigma$.
Unfortunately, this formulation fails 
to address the competition between classes,
which again compromises 
classification accuracy.

The C+1-way multi-class approach 
includes outliers as the C+1-th class,
whereas the inlier probability is
a 2-way softmax between 
the max-logit over inlier classes
and the outlier logit.
To account for class disbalance
we modulate the loss due to outliers
with $\lambda_\textrm{C+1}$.
Nevertheless, this loss affects 
inlier classification weights,
which may be harmful 
when the negatives are noisy
(as in our case).

Finally, the two-head approach 
complements the C-way classification head
with a head which directly predicts 
the outlier probability
as illustrated in Fig.\,\ref{fig:model}. 
The classification head is trained on inliers
while the outlier detection head is trained 
both on inliers and outliers.
The outlier detection head uses 
the standard cross entropy loss
modulated with hyper-parameter $\lambda_\textrm{TH}$. 
We combine the resulting prediction maps 
to obtain semantic segmentation 
into C+1 classes.
The outlier detection head overrides 
the classification head
whenever the outlier probability 
is greater than a threshold.
Thus, the classification head is unaffected 
by the negative data,
which provides hope to preserve 
the baseline semantic segmentation accuracy
even when training on 
extremely large negative datasets.

\subsection{Resistance to noisy outlier labels and sensitivity
to negative objects in positive context}

Training outlier detection on 
a diverse negative dataset
has to confront noise 
in negative training data.
For example, our negative dataset, ImageNet-1k, 
contains several classes (e.g.\ cab, streetcar)
which are part of the Cityscapes ontology.
Additionally, several stuff 
classes from Cityscapes
(e.g.\ building, terrain) 
occur in ImageNet-1k backgrounds.
Nevertheless, these pixels 
are vastly outnumbered by true inliers.
This is especially the case
when the training only considers 
the bounding box of the object 
which defines the ImageNet-1k class.

We address this issue 
by training our models 
on mixed batches 
with approximately equal share 
of inlier and negative images.
Thus, we perform many inlier epochs
during one negative epoch,
since our negative training dataset 
is much larger than the inlier ones. 
The proposed batch formation procedure
prevents occasional inliers from negative images
to significantly affect the training.
Additionally, it also favours 
stable development of batchnorm statistics.
Hence, the proposed training approach 
stands a fair chance to succeed.

We promote detection 
of outlier objects in inlier context
by pasting negative content 
into inlier training images.
We first resize the negative image 
to 5\% of the inlier image,
and then paste it at random.
We perform this before the cropping,
so some crops may contain only inlier pixels.
Unlike \cite{blum19arxiv},
we do not use the Cityscapes ignore class 
since it contains many inliers.

\section{Experiments}

We train most models on  
Vistas inliers \cite{neuhold17iccv}
by mapping original labels 
to Citysca\-pes \cite{cordts15cvpr} classes.
In some experiments we also use 
Cityscapes inliers 
to improve results
and explore influence of the domain shift. 
We train all applicable models on outliers 
from two variants of ImageNet-1k: 
the full dataset (ImageNet-1k-full) 
and the subset in which
bounding box annotations
are available (ImageNet-1k-bb).
In the latter case
we use only the bounding box 
for training on negative images 
(the remaining pixels are ignored)
and pasting into positive images.

We validate semantic segmentation 
by measuring mIoU on WildDash val
separately from outlier detection. 
We validate outlier detection
by measuring pixel level 
average precision (AP) 
in two different setups:
i) entire images are either
negative or positive, and
ii) appearance of negative
objects in positive context.
The former setup consists of
many assays
across WildDash val and random
subsets of LSUN images.
The LSUN subsets are dimensioned so that
the numbers of pixels in LSUN and WildDash val
are approximately equal.
Our experiments report mean and standard deviation
of the detection AP across 50 assays.
The latter setup involves WildDash val images
with pasted Pascal animals.
We select animals which take up
at least 1\% of the WildDash resolution,
and paste them at random in each WildDash image.

We normalize all images 
with ImageNet mean and variance,
and resize them so that 
the shorter side is 512 pixels.
We form training batches 
with random 512$\times$512 crops
which we jitter with horizontal flipping.
We set the auxiliary loss weight to 0.4 
and the classifier loss weight to 0.6.
We set $\lambda_\textrm{KL}$=0.2, 
$\lambda_\textrm{C+1}$=0.05,
and $\lambda_\textrm{TH}$=0.2. 
We use the standard Adam optimizer 
and divide the learning rate 
of pretrained parameters by 4.
All our models are trained 
throughout 75 Vistas epochs,
which corresponds to 
2 epochs of ImageNet-1k-full,  
or 5 epochs of ImageNet-1k-bb.
We detect outliers by thresholding 
inlier probability at $p_\textrm{IP}=0.5$.
Our models produce a dense index map
for C Cityscapes classes and 1 void class.
We obtain predictions at 
the benchmark resolution 
by bilinear upsampling.

We perform ODIN inference as follows.
First, we perform the forward pass
and the backward pass with respect 
to the max-softmax activation
(we use temperature T=10).
Then we determine 
the max-softmax gradient 
with respect to pixels.
We determine the perturbation
by multiplying the sign of the gradient 
with $\varepsilon$=0.001. 
Finally, we perturb 
the normalized input image,
perform another forward pass
and detect outliers according 
to the max-softmax criterion.

\subsection{Evaluation on the WildDash Benchmark}
\label{ss:benchmark}



Table \ref{table:bench_results} 
presents our results on 
the WildDash semantic segmentation benchmark, and compares them to other submissions
with accompanying publications.

\setlength{\tabcolsep}{4pt}
\begin{table}[htb!]
\begin{center}
\caption{Evaluation of the semantic segmentation
        models on WildDash bench}
\label{table:bench_results}
\begin{tabular}{|c||c|cccc|c|}
  \hline
  & \multicolumn{1}{c|}{Meta Avg}  
  & \multicolumn{4}{c|}{Classic} 
  & \multicolumn{1}{c|}{Negative}
\\
  \cline{2-7}
  Model
  & mIoU
  & mIoU
  & iIoU
  & mIoU
  & iIoU
  & mIoU\\
 &cla &cla&cla &cat&cat&cla\\
 \hline
 \hline
  \multicolumn{1}{|l||}{APMoE\_seg\_ROB \cite{kong2018pag}} & 22.2 & 22.5 & 12.6 & 48.1 & 35.2 & 22.8\\
  \hline
  \multicolumn{1}{|l||}{DRN\_MPC \cite{Yu2017}} & 28.3 & 29.1 & 13.9 & 49.2 & 29.2 & 15.9 \\
  \hline
  \multicolumn{1}{|l||}{DeepLabv3+\_CS \cite{chen2018encoder}} & 30.6 & 34.2 & 24.6 & 49.0 & 38.6 & 15.7\\ 
  \hline
  \multicolumn{1}{|l||}{LDN2\_ROB \cite{kreso18arxiv}} & 32.1 & 34.4 & 30.7 & 56.6 & 47.6 & 29.9\\
  \hline
  \multicolumn{1}{|l||}{MapillaryAI\_ROB \cite{bulo2017place}} &38.9 & 41.3 & \textbf{38.0} & 60.5 & \textbf{57.6} & 25.0\\
  \hline
  \multicolumn{1}{|l||}{AHiSS\_ROB \cite{meletis2018training}} & 39.0 & 41.0 & 32.2 & 53.9 & 39.3 & 43.6\\
  \hline
  \hline
  \multicolumn{1}{|l||}{LDN\_BIN 
    (ours, two-head)} & 
    41.8 & \textbf{43.8} & 
    37.3 & 58.6 & 53.3 & \textbf{54.3}\\
  \hline
  \multicolumn{1}{|l||}{LDN\_OE 
    (ours, C$\times$ multi-class)} &
    \textbf{42.7} & 43.3 & 
    31.9 & \textbf{60.7} & 50.3 & 52.8\\
\hline
\end{tabular}
\end{center}
\end{table}

Our two models use the same backbone
(DenseNet-169 \cite{huang17cvpr})
and different outlier detectors.
The LDN\_OE model has 
a single C-way multi-class head.
The LDN\_BIN model has two heads
as shown in Figure \ref{fig:model}. 
Both models have been trained on
Vistas train, Cityscapes train, 
and WildDash val (inliers),
as well as on ImageNet-1k-bb 
with pasting (outliers).
Our models significantly outperform
all previous submissions on negative images,
while also achieving the highest 
meta average mIoU
(the principal benchmark metric)
and the highest mIoU for classic images.
We achieve the second-best iIoU score 
for classic images, which indicates 
underperformance on small objects. 
This is likely due to the fact that
we train and evaluate our models 
on half resolution images. 

\subsection{Validation of Dense 
  Outlier Detection Approaches}

Table \ref{table:OOD_detection} 
compares various approaches 
for dense outlier detection.
All models are based on DenseNet-169,
and trained on Vistas (inliers).
The first section shows the results 
of a C-way multi-class model 
trained without outliers,
where outliers are detected 
with max-softmax \cite{hendrycks17iclr}
and ODIN + max-softmax \cite{liang18iclr}.
We note that ODIN slightly improves 
the results across all experiments.

\begin{table}[htb]
\centering
\caption{Validation of dense 
  outlier detection approaches.
  WD denotes WildDash val.}
\label{table:OOD_detection}
\begin{tabular}{|c|c||c|c|c|}
  \hline
  Model & ImageNet &
    \multicolumn{1}{c|}{AP WD-LSUN} &
    \multicolumn{1}{c|}{AP WD-Pascal} &
    \multicolumn{1}{c|}{mIoU WD}\\
 \hline
 \hline
  \multicolumn{1}{|l|}{
    C$\times$ multi-class} &
    \ding{55} &$55.65 \pm 0.80$ & 6.01 & 49.07
    \\
  \hline
  \multicolumn{1}{|l|}{
    C$\times$ multi-class, ODIN} & 
    \ding{55} & $55.98 \pm 0.77$ & 6.92 & \textbf{49.77} 
    \\
 \hline
 \hline
  \multicolumn{1}{|l|}{
    C+1$\times$ multi-class}& 
    \ding{51} & 
    $98.92 \pm 0.06$ & 33.59 & 45.60
    \\
 \hline
  \multicolumn{1}{|l|}{
    C$\times$ multi-label} & 
    \ding{51} & $98.75 \pm 0.07$ & \textbf{57.31} & 42.72
    \\
 \hline
  \multicolumn{1}{|l|}{
    C$\times$ multi-class} 
    & \ding{51} & $ \textbf{99.49} \pm \textbf{0.04}$ & 41.72 & 46.69 
    \\
 \hline
  \multicolumn{1}{|l|}{two heads} 
    & \ding{51} & $99.25 \pm 0.04$ & 46.83 & 47.37 
    \\
\hline
\end{tabular}
\end{table}

The second section of the table shows 
the four dense outlier detection approaches
(cf.\ Section \ref{ss:model})
which we also train on 
ImageNet-1k-bb with pasting (outliers).
Columns 3 and 4 clearly show
that training with noisy and diverse negatives
significantly improves outlier detection. 
However, we also note a reduction 
of the segmentation score 
as shown in the column 5.
This reduction is lowest for 
the C-way multi-class model 
and the two-head model,
which we analyze next.

The two-head model is slightly worse
in discriminating WildDash val from LSUN,
which indicates that it is
more sensitive to domain shift
between Vistas train and WildDash val.
On the other hand, the two-head model achieves 
better inlier segmentation
(0.7 pp, column 5),
and much better outlier detection
on Pascal animals (5 pp, column 4).
A closer inspection shows 
that these advantages occur
since the single-head C-way approach
generates many false positive
outlier detections at semantic borders
due to lower max-softmax.

The C+1-way multi-class model performs the worst 
out of all models trained with noisy outliers.
%
The sigmoid model performs 
well on outlier detection
but underperforms on inlier segmentation.

\subsection{Validation of Dense 
  Feature Extractor Backbones}

Table \ref{table:backbone_val} explores 
influence of different backbones
to the performance of our two-head model. 
We experiment with ResNets and 
DenseNets of varying depths.
The upsampling blocks are connected 
with the first three DenseNet blocks,
as shown in Fig.\,\ref{fig:model}.
In the ResNet case, the upsampling blocks 
are connected with the last addition
at the corresponding subsampling level.
We train on Vistas (inliers) 
and ImageNet-1k-bb with pasting (outliers).

\begin{table}[htb]
\centering
\caption{Validation of backbones
  for the two-head model.
  WD denotes WildDash val.
  }
\label{table:backbone_val}
\begin{tabular}{|c||c|c|c|}
  \hline
  Backbone & 
    \multicolumn{1}{c|}{AP WD-LSUN } & 
    \multicolumn{1}{c|}{AP WD-Pascal} &
    \multicolumn{1}{c|}{mIoU WD}\\
 \hline
 \hline
  \multicolumn{1}{|l||}{DenseNet-121} & $99.05 \pm 0.03$ & 55.84 & 44.75 \\
 \hline
  \multicolumn{1}{|l||}{DenseNet-169} & $\textbf{99.25} \pm \textbf{0.04}$ & 46.83 & 47.37\\
 \hline
  \multicolumn{1}{|l||}{DenseNet-201} & $98.34 \pm 0.07$ &	36.88 &	\textbf{47.59} \\
 \hline
  \multicolumn{1}{|l||}{ResNet-34} & $97.19 \pm 0.07$ & 47.24 & 45.17\\
 \hline
  \multicolumn{1}{|l||}{ResNet-50} & $99.10 \pm 0.04$ & \textbf{56.18} & 41.65 \\
 \hline
  \multicolumn{1}{|l||}{ResNet-101} & $98.96 \pm 0.06$ & 52.02 & 43.67 \\
\hline
\end{tabular}
\end{table}

All models achieve very good 
outlier detection in negative images.
There appears to be a trade-off between
detection of outliers at negative objects
and semantic segmentation accuracy.
We opt for better semantic segmentation results
since WildDash test does not have negative objects
in positive context.
We therefore use the DenseNet-169 backbone 
in most other experiments due to 
a very good overall performance.

\subsection{Influence of the Training Data}

Table \ref{table:dataset_results_in}
explores the influence of 
inlier training data 
to the model performance.
All experiments involve the two-head model 
based on DenseNet-169, which was trained on outliers from ImageNet-1k-bb with pasting.

\begin{table}[htb]
\centering
\caption{Influence of
  the inlier training dataset 
  to the performance of the two-head model 
  with the DenseNet-169 backbone.
  WD denotes WildDash val.
}
\label{table:dataset_results_in}
\begin{tabular}{|c||c|c|c|}
  \hline
  Inlier training dataset &
  AP WD-LSUN  & 
  AP WD-Pascal &
  mIoU WD\\
 \hline
  \hline
  \multicolumn{1}{|l||}{Cityscapes} & 
    $66.57 \pm 0.86$ & 13.85 & 11.12\\
 \hline
  \multicolumn{1}{|l||}{Vistas} & 
    $99.25 \pm 0.04$ & 46.83 & 47.17 \\
 \hline
  \multicolumn{1}{|l||}{Cityscapes, Vistas} &
    $\textbf{99.29} \pm \textbf{0.03}$ & \textbf{53.68} & \textbf{47.78} \\
\hline
\end{tabular}
\end{table}

The results suggest that there is 
a very large domain shift between 
Cityscapes and WildDash val.
Training on inliers from Cityscapes 
leads to very low AP scores,
which indicates that many WildDash val pixels
are predicted as outliers
with respect to Cityscapes.
This suggests that Cityscapes 
is not an appropriate training dataset 
for real-world applications.
Training on inliers from Vistas 
leads to much better results
which is likely due to greater variety 
with respect to camera, time of day, 
weather, resolution etc. 
The best results across the board 
have been achieved
when both inlier datasets
are used for training.

Table \ref{table:dataset_results_out} 
explores the impact of  
negative training data.
All experiments feature 
the two-head model with DenseNet-169
trained on inliers from Vistas.

\begin{table}[htb]
\centering
\caption{Influence of the 
  outlier training dataset 
  to the performance of our two-head model 
  with the DenseNet-169 backbone. 
  WD denotes WildDash val.
}
\label{table:dataset_results_out}
\begin{tabular}{|l|c||c|c|}
  \hline
  \multirow{1}*{Outlier training dataset} &
    outlier pasting &
    \multicolumn{1}{c|}{AP WD-Pascal} &
    \multicolumn{1}{c|}{mIoU WD}\\
 \hline
  \hline
  \multicolumn{1}{|l|}{ImageNet-1k-full} &
    no & 2.94 &  43.13\\
 \hline
  \multicolumn{1}{|l|}{ImageNet-1k-full} &
    yes & 45.96 & 43.68\\
 \hline
  \multicolumn{1}{|l|}{ImageNet-1k-bb} &
    yes & \textbf{46.83}  & \textbf{47.17} \\
\hline
\end{tabular}
\end{table}
The table shows that training 
with pasted negatives greatly improves
outlier detection on negative objects.
It is intuitively clear that
a model which never sees a border 
between inliers and outliers 
during training 
does not stand a chance
to accurately locate such borders 
during inference.

The table also shows that ImageNet-1k-bb
significantly boosts inlier segmentation, 
while also improving outlier detection
on negative objects.
We believe that this occurs because
ImageNet-1k-bb has a smaller overlap 
with respect to the inlier training data,
due to high incidence of Cityscapes classes
(e.g. vegetation, sky, road) 
in ImageNet backgrounds.
This simplifies outlier detection 
due to decreased noise in the training set,
and allows more capacity of
the shared feature extractor
to be used for the segmentation task.
The table omits outlier detection 
in negative images, 
since all models 
achieve over 99 \% AP
on that task.

%

%
%

\subsection{Comparing the Two-Head and 
  C-way Multi-class Models}
\label{ss:comparison}

We now compare our two models
from Table \ref{table:bench_results}
in more detail.
We remind that the two models 
have the same feature extractor
and are trained on the same data.
The two-head model performs better 
in most classic evaluation categories 
as well as in the negative category, 
however it has a lower meta average score.

Table \ref{table:bin_oe_hazard} explores 
influence of WildDash hazards
\cite{zendel18eccv}
on the performance of the two models. 
The C-way multi-class model
has a lower performance drop 
in most hazard categories. 
The difference is especially large
in images with distortion and overexposure. 
Qualitative experiments show 
that this occurs since 
the two-head model tends 
to recognize pixels in 
images with hazards as outliers
(cf.\ Fig.\,\ref{fig:bench_bin_oe}).

\begin{table}[htb]
\begin{center}
\caption{Impact of hazards
  to performance of our 
  WildDash submissions.
  The hazards are
  image blur,
  uncommon road coverage,
  lens distortion,
  large ego-hood,
  occlusion,
  overexposure, 
  particles,
  dirty windscreen,
  underexposure, and 
  uncommon variations.
  LDN\_BIN denotes 
  the two-head model.
  LDN\_OE denotes
  the C-way 
  multi-class model.
  }
\label{table:bin_oe_hazard}
\begin{tabular}{|c||cccccccccccc|}
  \hline
  \multirow{2}*{Model}& \multicolumn{10}{c|}{
    Class mIoU drop across WildDash hazards
    \cite{zendel18eccv}}
\\
  \cline{2-11}
  &blur
  &cov.
  &dist.
  &hood
  &occ.
  &over.
  &part.
  &screen
  &under.
  &\multicolumn{1}{c|}{var.}\\
 \hline
 \hline
  \multicolumn{1}{|l||}{\textsc{ldn\_bin}} &-14\%&-14\%&-22\%&-14\%&\textbf{-3\%}&-35\%&-3\%&-9\%&\textbf{-25\%}&\multicolumn{1}{c|}{-8\%}\\
  \hline
  \multicolumn{1}{|l||}{\textsc{ldn\_oe}} & \textbf{-11\%}&\textbf{-13\%}&\textbf{-7\%}&\textbf{-10\%}&-5\%&\textbf{-24\%}& \textbf{0\%}&\textbf{-6\%}&-30\%&\multicolumn{1}{c|}{\textbf{-7\%}}\\
  \hline

\end{tabular}
\end{center}
\end{table}

Fig.\,\ref{fig:bench_bin_oe} presents
a qualitative comparison of our two 
submissions to the WildDash benchmark.
Experiments in rows 1 and 2 show
that the two-head model 
performs better in classic images
due to better performance on semantic borders.
Furthermore, the two-head model is also better
in detecting negative objects in positive context 
(ego-vehicle, the forklift, and the horse). 
Experiments in row 3 show 
that the two-head model 
tends to recognize all pixels 
in images with overexposure 
and distortion hazards as outliers.
Experiments in rows 4 and 5 show that
the two-head model recognizes 
entire negative images as outliers,
while the C-way single-head model
is able to recognize positive objects
(the four persons) in negative images.
These experiments suggest
that both models are able to detect
outliers at visual classes 
which are (at least nominally)
not present in ImageNet-1k:
ego-vehicle, toy-brick construction,
digital noise, and text.
Fig.\,\ref{fig:bench2_bin_oe} 
illustrates space for further improvement
by presenting a few
failure cases on WildDash test.

\newcommand{\mylen}{0.49\textwidth}
\begin{figure}[htb]
  \centering
  \includegraphics[width=\mylen]{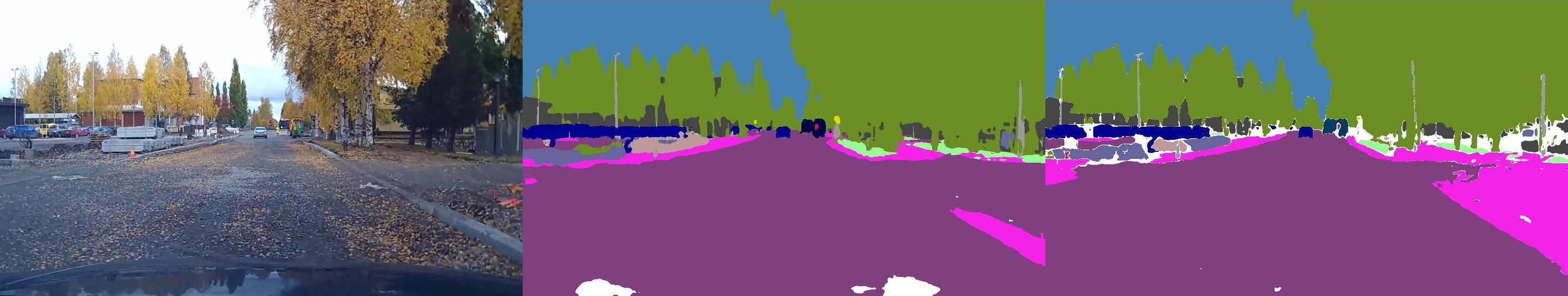}
  \hfill
  \includegraphics[width=\mylen]{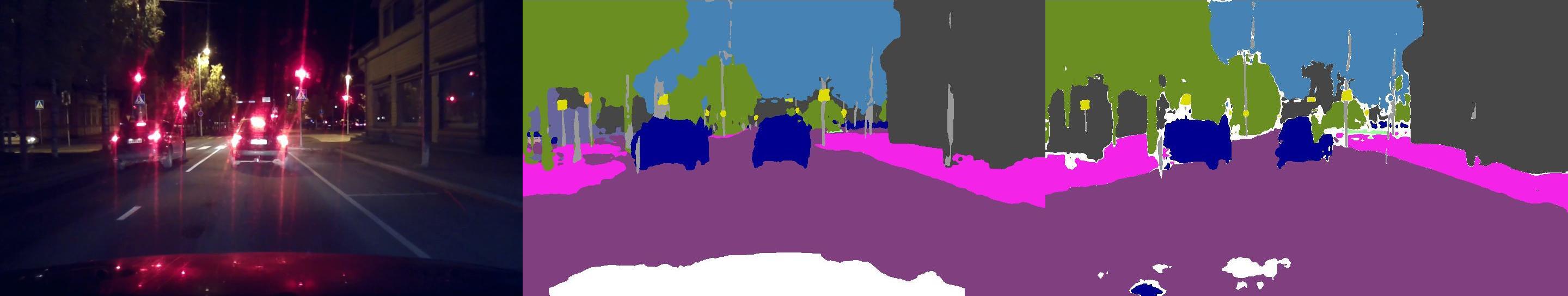}
  \\[0.2em]
  \includegraphics[width=\mylen]{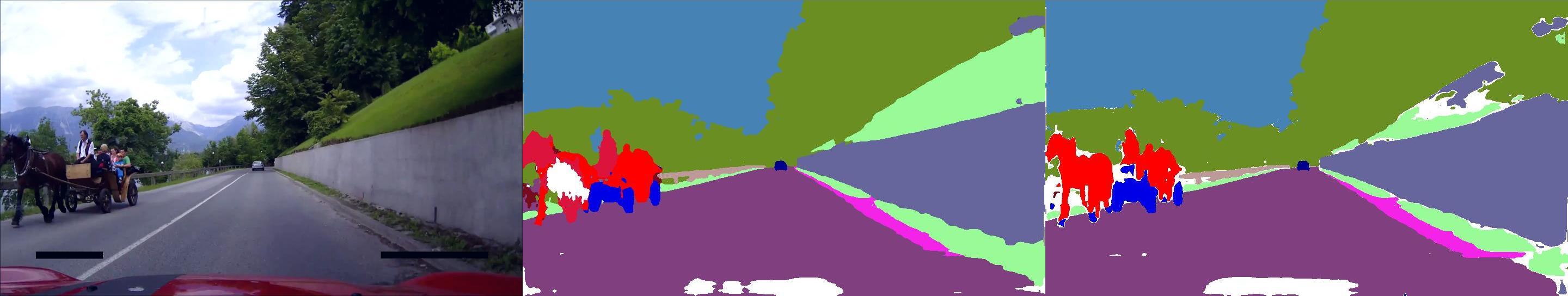}
  \hfill
  \includegraphics[width=\mylen]{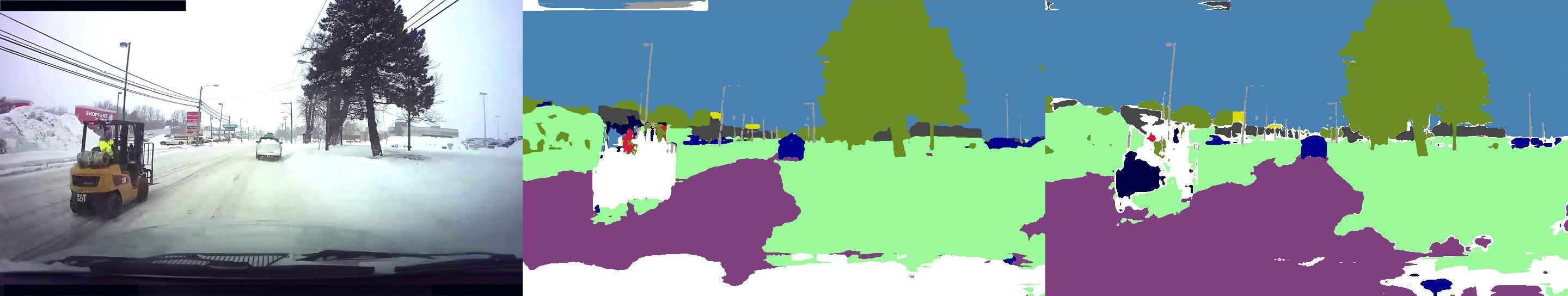}
  \\[0.2em]
  \includegraphics[width=\mylen]{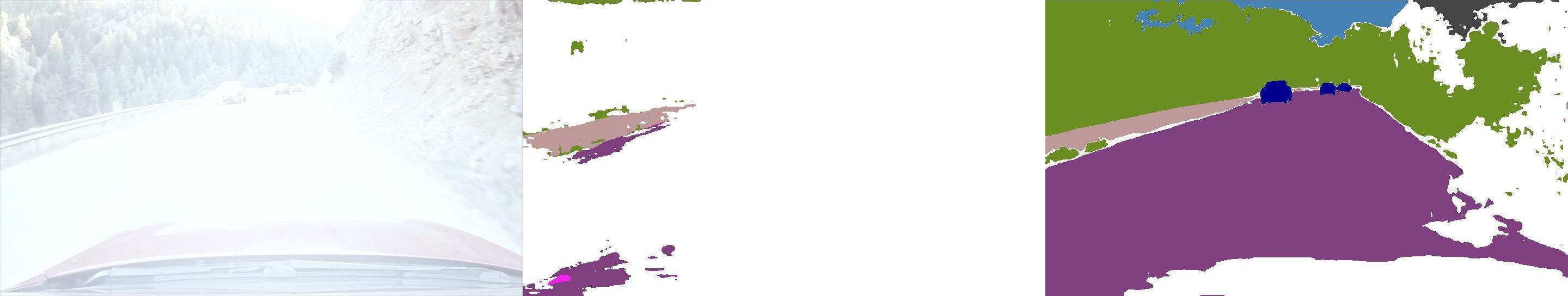}
  \hfill
  \includegraphics[width=\mylen]{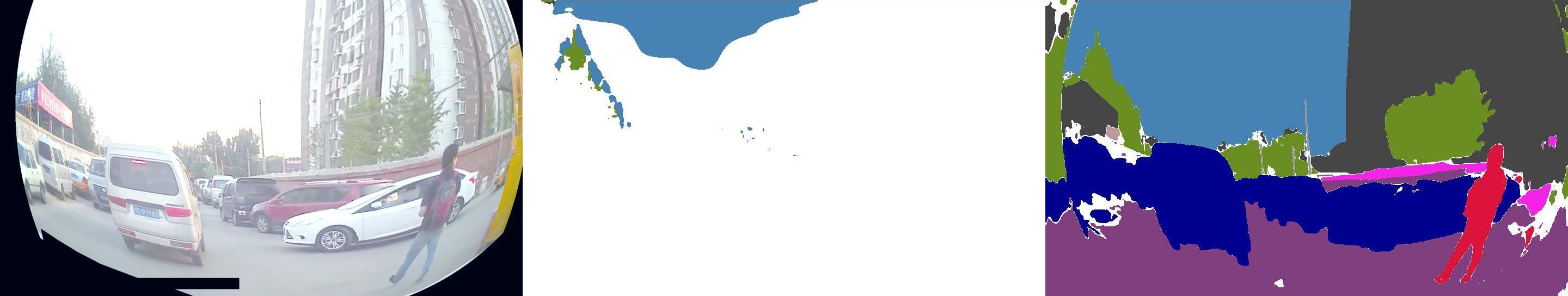}
  \\[0.2em]
  \includegraphics[width=\mylen]{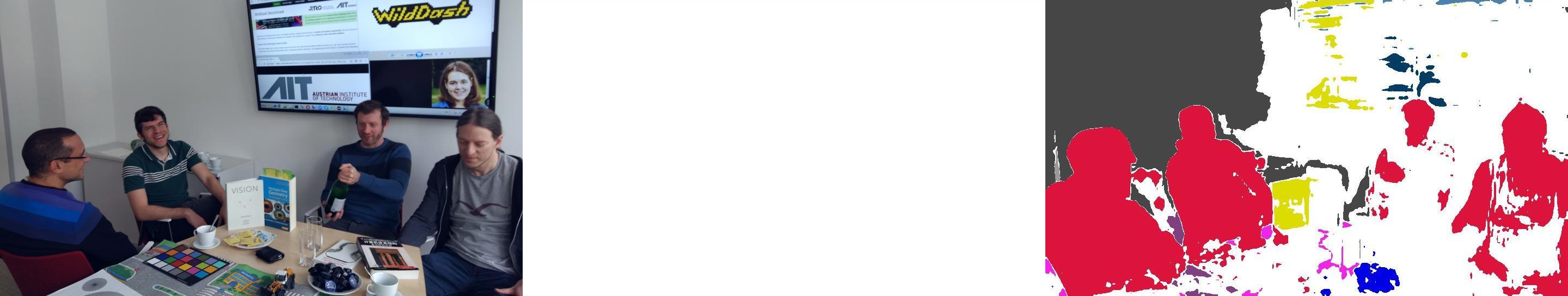}
  \hfill
  \includegraphics[width=\mylen]{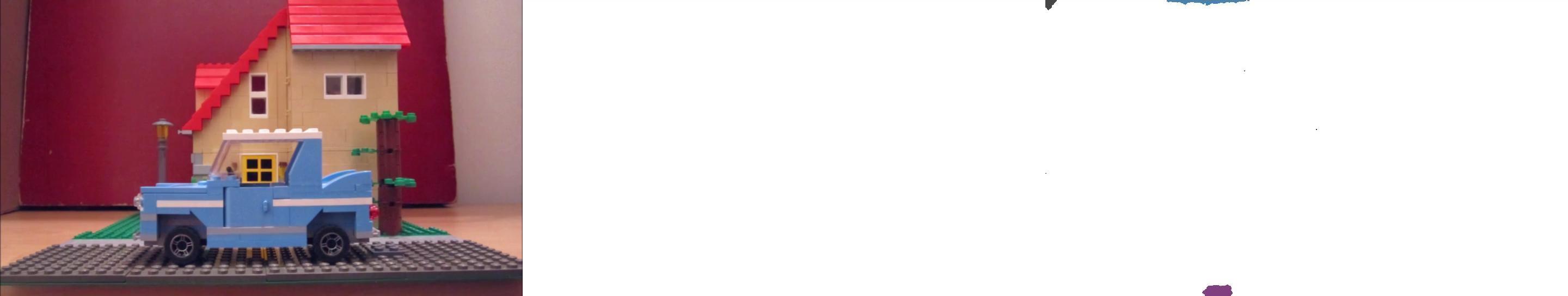}
  \\[0.2em]
  \includegraphics[width=\mylen]{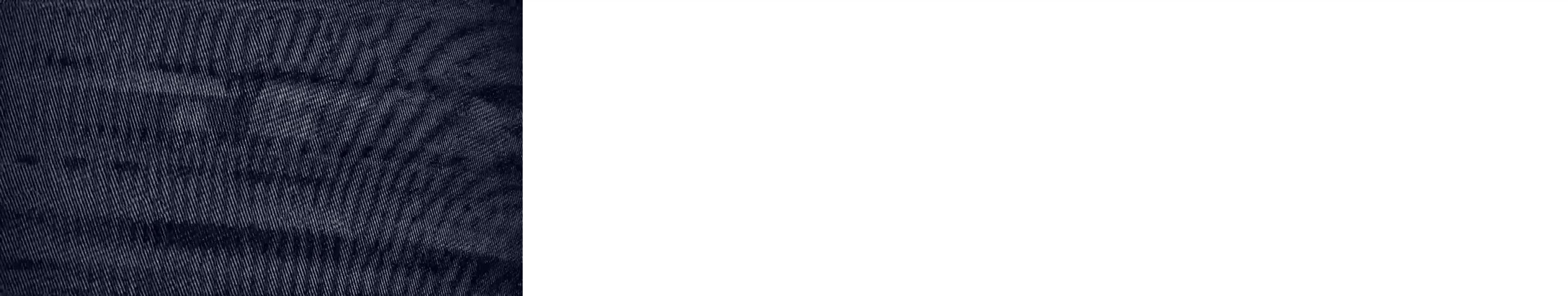}
  \hfill
  \includegraphics[width=\mylen]{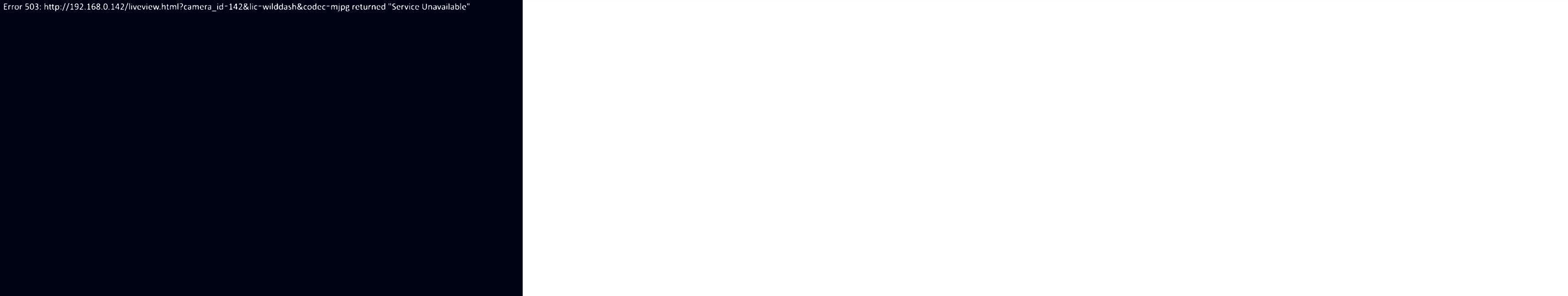}
  \caption{Qualitative performance 
  of our two submissions 
  to the WildDash benchmark. 
  Each triplet contains 
  a test image (left),
  the output of the two-head model (center), 
  and the output of the model trained 
  to predict uniform distribution in outliers (right).
  Rows represent inlier images (1),
  outlier objects in inlier context (2), 
  inlier images with hazards (3),
  out-of-scope negatives (4),
  and abstract negatives (5).
  The two-head model produces 
  more outlier detections
  while performing better 
  in classic images 
  (cf.~Table\,\ref{table:bench_results}).
  }
  \label{fig:bench_bin_oe}
\end{figure}

\begin{figure}[htb]
  \centering
  \includegraphics[width=\mylen]{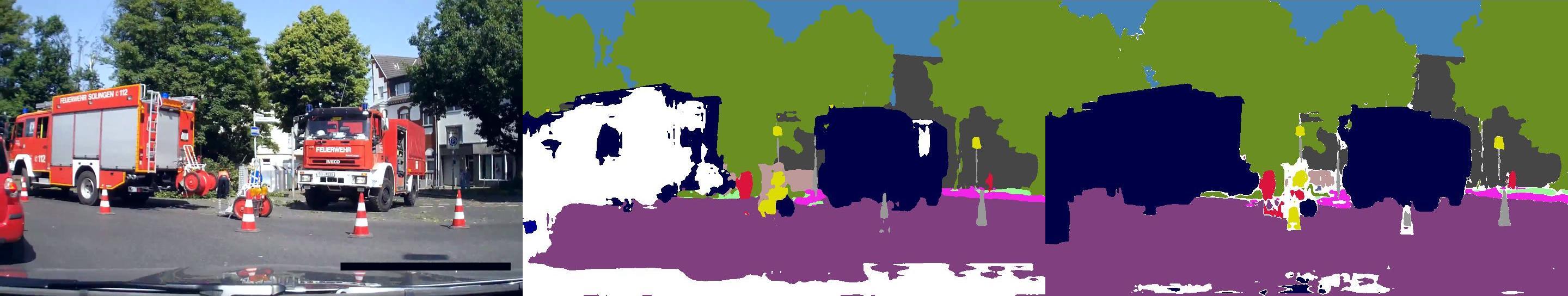}
  \hfill
  \includegraphics[width=\mylen]{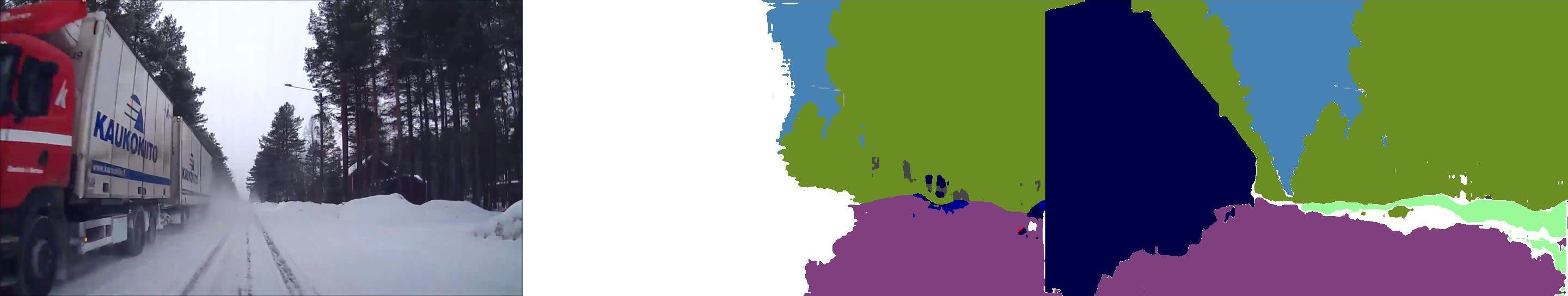}
  \\[0.2em]
  \includegraphics[width=\mylen]{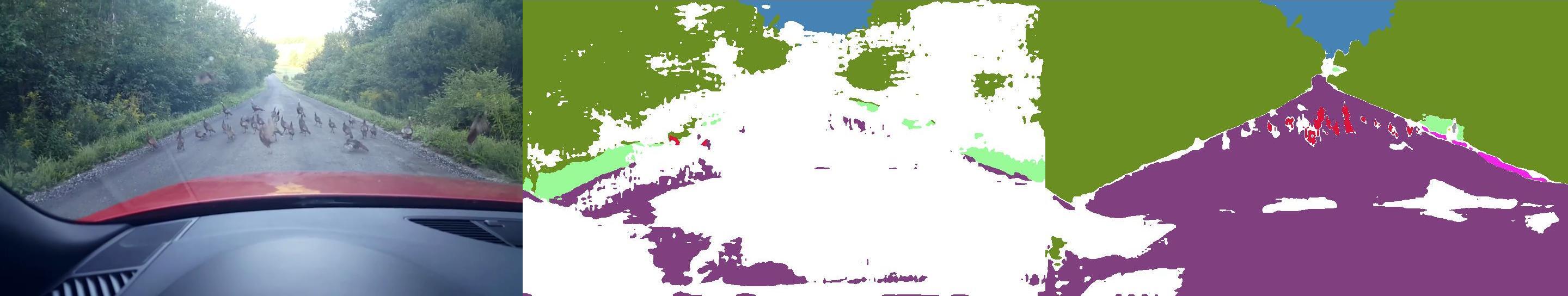}
  \hfill
  \includegraphics[width=\mylen]{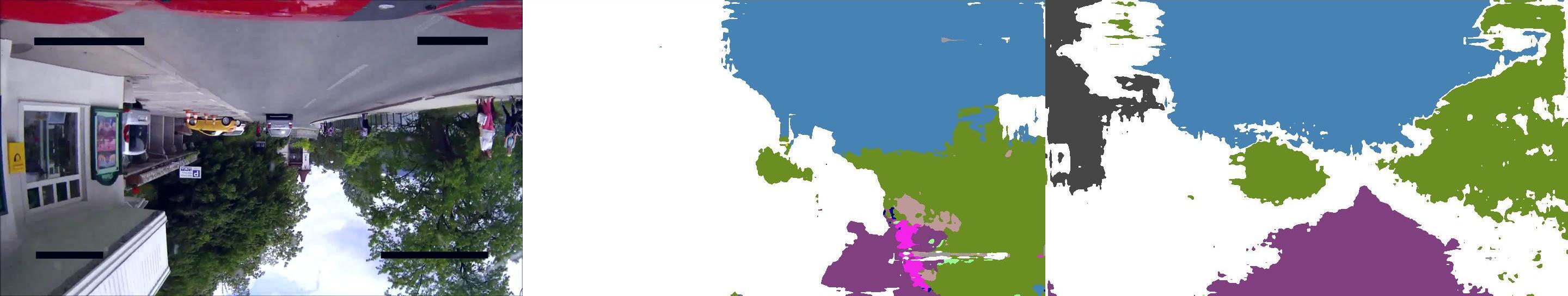}
  \caption{%
    Failure cases on WD test
    arranged as in 
    Fig.\,\ref{fig:bench_bin_oe}.
    Top: 
    the two-head model predicts 
    outliers at trucks.
    Bottom left:
    both models 
    fail to accurately detect 
    birds on the road.
    Bottom right:
    both models exchange road and sky
    due to position bias
    (late experiments show this 
     can be improved with training 
     on smaller crops and scale jittering).
  }
  \label{fig:bench2_bin_oe}
\end{figure}
\subsection{Discussion}

Validation on LSUN and WildDash val suggests 
that detecting entire outlier images 
is an easy problem.
A useful practical application of this result
would be a module to detect 
whether the camera is out of order
due to being covered, dirty or faulty.
Validation on pasted Pascal animals 
suggests that detecting outliers
in inlier context
is somewhat harder but still 
within reach of modern techniques.

Evaluation on WildDash test shows 
that our models outperform 
previous published approaches.
We note significant improvement
with respect to state of the art
in classic and negative images,
as well as in average mIoU score.
Our models successfully detect all 
abstract and out-of-scope negatives \cite{zendel18eccv},
even though much of this content 
is not represented by ImageNet-1k classes.

Future work should address further development
of open-set evaluation data\-sets 
such as \cite{zendel18eccv,blum19arxiv}.
In particular, the community would benefit
from substantially larger negative test sets
which should include diverse non-ImageNet-1k content,
as well as outlier objects in inlier context 
and inlier objects in outlier context.

\section{Conclusion}


We have presented an approach to combine 
semantic segmentation and 
dense outlier detection
without significantly deteriorating 
either of the two tasks. 
We cast outlier detection 
as binary classification 
on top of a shared convolutional re\-presentation.
This allows for solving both tasks 
with a single forward pass 
through a convolutional backbone.
We train on inliers from
standard road driving datasets 
(Vista, Cityscapes),
and noisy outliers from a very diverse 
negative dataset (ImageNet-1k).
The proposed training procedure tolerates 
inliers in negative training images
and generalizes
to images with mixed content 
(inlier background, outlier objects).
We perform extensive open-set validation
on WildDash val (inliers),
LSUN val (outliers),
and pasted Pascal objects (outliers).
The results confirm suitability
of the proposed training procedure.
The proposed multi-head model outperforms 
the C-way multi-label model 
and the C+1-way multi-class model,
while performing comparably to the 
C-way multi-class model trained 
to predict uniform distribution in outliers.
We apply our two best models to WildDash test 
and set a new state of the art
on the WildDash benchmark.

\bibliographystyle{splncs04}
\bibliography{egbib}

\begin{thebibliography}{10}
\providecommand{\url}[1]{\texttt{#1}}
\providecommand{\urlprefix}{URL }
\providecommand{\doi}[1]{https://doi.org/#1}

\bibitem{bengio13pami}
Bengio, Y., Courville, A.C., Vincent, P.: Representation learning: {A} review
  and new perspectives. {IEEE} Trans. Pattern Anal. Mach. Intell.
  \textbf{35}(8),  1798--1828 (2013)

\bibitem{bevandic2018}
Bevandic, P., Kreso, I., Orsic, M., Segvic, S.: Discriminative
  out-of-distribution detection for semantic segmentation. CoRR
  \textbf{abs/1808.07703} (2018)

\bibitem{blum19arxiv}
Blum, H., Sarlin, P., Nieto, J.I., Siegwart, R., Cadena, C.: The {F}ishyscapes
  benchmark: Measuring blind spots in semantic segmentation. CoRR
  \textbf{abs/1904.03215}

\bibitem{brock19iclr}
Brock, A., Donahue, J., Simonyan, K.: Large scale {GAN} training for high
  fidelity natural image synthesis. In: ICLR (2019)

\bibitem{brostow08eccv}
Brostow, G.J., Shotton, J., Fauqueur, J., Cipolla, R.: Segmentation and
  recognition using structure from motion point clouds. In: ECCV. pp. 44--57
  (2008)

\bibitem{bulo2017place}
Bul{\`o}, S.R., Porzi, L., Kontschieder, P.: In-place activated batchnorm for
  memory-optimized training of dnns. CoRR, abs/1712.02616, December  \textbf{5}
  (2017)

\bibitem{Caruana1997}
Caruana, R.: Multitask learning. Machine Learning  \textbf{28}(1),  41--75 (Jul
  1997). \doi{10.1023/A:1007379606734}

\bibitem{chen2018encoder}
Chen, L.C., Zhu, Y., Papandreou, G., Schroff, F., Adam, H.: Encoder-decoder
  with atrous separable convolution for semantic image segmentation. In: ECCV
  (2018)

\bibitem{cordts15cvpr}
Cordts, M., Omran, M., Ramos, S., Scharw{\"a}chter, T., Enzweiler, M.,
  Benenson, R., Franke, U., Roth, S., Schiele, B.: The cityscapes dataset. In:
  CVPRW (2015)

\bibitem{deng09cvpr}
Deng, J., Dong, W., Socher, R., Li, L., Li, K., Li, F.: Imagenet: {A}
  large-scale hierarchical image database. In: CVPR. pp. 248--255 (2009)

\bibitem{devries18arxiv}
DeVries, T., Taylor, G.W.: Learning confidence for out-of-distribution
  detection in neural networks. CoRR  \textbf{abs/1802.04865} (2018)

\bibitem{Eigen2015PredictingDS}
Eigen, D., Fergus, R.: Predicting depth, surface normals and semantic labels
  with a common multi-scale convolutional architecture. ICCV pp. 2650--2658
  (2015)

\bibitem{everingham10ijcv}
Everingham, M., Gool, L., Williams, C.K., Winn, J., Zisserman, A.: The pascal
  visual object classes (voc) challenge. Int. J. Comput. Vision  (2010)

\bibitem{geiger13ijrr}
Geiger, A., Lenz, P., Stiller, C., Urtasun, R.: Vision meets robotics: The
  kitti dataset. International Journal of Robotics Research (IJRR)  (2013)

\bibitem{goodfellow14nips}
Goodfellow, I.J., Pouget-Abadie, J., Mirza, M., Xu, B., Warde-Farley, D.,
  Ozair, S., Courville, A.C., Bengio, Y.: Generative adversarial nets. In: NIPS
  (2014)

\bibitem{guo17icml}
Guo, C., Pleiss, G., Sun, Y., Weinberger, K.Q.: On calibration of modern neural
  networks. In: ICML. pp. 1321--1330 (2017)

\bibitem{mask}
{He}, K., {Gkioxari}, G., {Dollár}, P., {Girshick}, R.: Mask {R-CNN}. In: ICCV
  (2017)

\bibitem{he14eccv}
He, K., Zhang, X., Ren, S., Sun, J.: Spatial pyramid pooling in deep
  convolutional networks for visual recognition. In: ECCV. pp. 346--361 (2014)

\bibitem{He2016DeepRL}
He, K., Zhang, X., Ren, S., Sun, J.: Deep residual learning for image
  recognition. CVPR pp. 770--778 (2016)

\bibitem{hendrycks17iclr}
Hendrycks, D., Gimpel, K.: A baseline for detecting misclassified and
  out-of-distribution examples in neural networks. In: ICLR (2017)

\bibitem{hendrycks19iclr}
Hendrycks, D., Mazeika, M., Dietterich, T.: Deep anomaly detection with outlier
  exposure. In: ICLR (2019)

\bibitem{huang17cvpr}
Huang, G., Liu, Z., Weinberger, K.Q.: Densely connected convolutional networks.
  In: CVPR (2017)

\bibitem{kendall15arxiv}
Kendall, A., Badrinarayanan, V., Cipolla, R.: Bayesian segnet: Model
  uncertainty in deep convolutional encoder-decoder architectures for scene
  understanding. CoRR  \textbf{abs/1511.02680} (2015)

\bibitem{kendall17nips}
Kendall, A., Gal, Y.: What uncertainties do we need in bayesian deep learning
  for computer vision? In: NIPS. pp. 5574--5584 (2017)

\bibitem{kong2018pag}
Kong, S., Fowlkes, C.: Pixel-wise attentional gating for parsimonious pixel
  labeling. In: arxiv 1805.01556 (2018)

\bibitem{kreso17cvrsuad}
Kreso, I., Krapac, J., Segvic, S.: Ladder-style densenets for semantic
  segmentation of large natural images. In: ICCV CVRSUAD 2017. pp. 238--245
  (2017)

\bibitem{kreso19arxiv}
Kreso, I., Krapac, J., Segvic, S.: Efficient ladder-style densenets for
  semantic segmentation of large images. CoRR  \textbf{abs/1905.05661} (2019)

\bibitem{kreso18arxiv}
Kreso, I., Orsic, M., Bevandic, P., Segvic, S.: Robust semantic segmentation
  with ladder-densenet models. CoRR  \textbf{abs/1806.03465} (2018)

\bibitem{lakshminarayanan17nips}
Lakshminarayanan, B., Pritzel, A., Blundell, C.: Simple and scalable predictive
  uncertainty estimation using deep ensembles. In: NIPS. pp. 6402--6413 (2017)

\bibitem{lee18iclr}
Lee, K., Lee, H., Lee, K., Shin, J.: Training confidence-calibrated classifiers
  for detecting out-of-distribution samples. In: ICLR (2018)

\bibitem{Lee2018ASU}
Lee, K., Lee, K., Lee, H., Shin, J.: A simple unified framework for detecting
  out-of-distribution samples and adversarial attacks. In: NeurIPS (2018)

\bibitem{liang18iclr}
Liang, S., Li, Y., Srikant, R.: Enhancing the reliability of
  out-of-distribution image detection in neural networks. In: ICLR (2018)

\bibitem{lintsungyi17cvpr}
Lin, T., Doll{\'{a}}r, P., Girshick, R.B., He, K., Hariharan, B., Belongie,
  S.J.: Feature pyramid networks for object detection. In: CVPR. pp. 936--944
  (2017)

\bibitem{meletis2018training}
{Meletis}, P., {Dubbelman}, G.: Training of convolutional networks on multiple
  heterogeneous datasets for street scene semantic segmentation. In: IV (2018)

\bibitem{nalisnick19iclr}
Nalisnick, E.T., Matsukawa, A., Teh, Y.W., G{\"{o}}r{\"{u}}r, D.,
  Lakshminarayanan, B.: Do deep generative models know what they don't know?
  In: ICLR (2019)

\bibitem{neuhold17iccv}
Neuhold, G., Ollmann, T., Bul{\`{o}}, S.R., Kontschieder, P.: The mapillary
  vistas dataset for semantic understanding of street scenes. In: ICCV (2017)

\bibitem{ngiam2011}
Ngiam, J., Khosla, A., Kim, M., Nam, J., Lee, H., Y.~Ng, A.: Multimodal deep
  learning. In: ICML. pp. 689--696 (2011)

\bibitem{sabokrou2018adversarially}
Sabokrou, M., Khalooei, M., Fathy, M., Adeli, E.: Adversarially learned
  one-class classifier for novelty detection. In: CVPR. pp. 3379--3388 (2018)

\bibitem{scheirer13pami}
Scheirer, W.J., de~Rezende~Rocha, A., Sapkota, A., Boult, T.E.: Toward open set
  recognition. {IEEE} Trans. Pattern Anal. Mach. Intell.  \textbf{35}(7),
  1757--1772 (2013)

\bibitem{shafaei2018}
Shafaei, A., Schmidt, M., Little, J.J.: Does your model know the digit 6 is not
  a cat? {A} less biased evaluation of "outlier" detectors  (2018)

\bibitem{smith18uai}
Smith, L., Gal, Y.: Understanding measures of uncertainty for adversarial
  example detection. In: UAI. vol. abs/1803.08533 (2018)

\bibitem{torralba2011}
{Torralba}, A., {Efros}, A.A.: Unbiased look at dataset bias. In: CVPR (June
  2011). \doi{10.1109/CVPR.2011.5995347}

\bibitem{Vyas2018OutofDistributionDU}
Vyas, A., Jammalamadaka, N., Zhu, X., Das, D., Kaul, B., Willke, T.L.:
  Out-of-distribution detection using an ensemble of self supervised leave-out
  classifiers. In: ECCV (2018)

\bibitem{Yu2017}
Yu, F., Koltun, V., Funkhouser, T.: Dilated residual networks. In: CVPR (2017)

\bibitem{yu2015}
Yu, F., Zhang, Y., Song, S., Seff, A., Xiao, J.: {LSUN:} construction of a
  large-scale image dataset using deep learning with humans in the loop  (2015)

\bibitem{zamir18cvpr}
Zamir, A.R., Sax, A., Shen, W.B., Guibas, L.J., Malik, J., Savarese, S.:
  Taskonomy: Disentangling task transfer learning. In: CVPR (2018)

\bibitem{zendel18eccv}
Zendel, O., Honauer, K., Murschitz, M., Steininger, D., Fernandez~Dominguez,
  G.: Wilddash - creating hazard-aware benchmarks. In: ECCV (September 2018)

\bibitem{zendel17ijcv}
Zendel, O., Murschitz, M., Humenberger, M., Herzner, W.: How good is my test
  data? introducing safety analysis for computer vision. International Journal
  of Computer Vision  \textbf{125}(1-3),  95--109 (2017)

\bibitem{zhao17cvpr}
Zhao, H., Shi, J., Qi, X., Wang, X., Jia, J.: Pyramid scene parsing network.
  In: CVPR (2017)

\end{thebibliography}

\setcounter{section}{0}
\renewcommand{\thesection}{\appendixname~\Alph{section}}%
\section{Supplementary material}
We use this supplement to further discuss the 
experiments from Section 4 of the main paper.
We clarify the losses used
for training the models and
expand on the quantitative results
described in the main paper by offering 
qualitative analysis of model outputs.
Furthermore, we offer experiments
with two additional outlier
detection approaches applied
to dense prediction: MC-dropout and trainable
confidence. 

In the first section we
provide a detailed overview of the
training losses.
The results of the additional experiments 
can be seen at the beginning of the second section.
The rest of the second and all of the third
section are dedicated to illustrating
validation experiments on WD-Pascal dataset.
WD-Pascal is a set of images 
which we create by pasting instances of
animals from PASCAL VOC 2007 dataset 
into WildDash validation images. 
We show that low AP scores on WD-Pascal 
are strongly affected by 
false positive detections in WildDash images
which occur due to 
sensitivity to domain shift.

In the fourth section we give more
examples of the performance of our
submissions to the WildDash benchmark.


\renewcommand{\thesection}{\Alph{section}}%

\subsection{Review of loss functions}
Segmentation can be viewed as multi-class
classification. To train it, we use the negative
log-likelihood
loss for each pixel of the output. 
We use the softmax activation function
on the output of the network. Additionally, as
a class balancing techinque, 
the loss for each pixel can be weighted depending
on the ground truth label. In our experiments
we only used class balancing during the
C+1 way multi-class training by setting
the $\lambda_{\mathrm{C+1}}$ to 0.05 to account
for the fact that outlier pixels
outnumber inlier pixels for
each individual class. In the case of
the model with trainable confidence,
the prediction is adjusted 
by interpolating between the original predictions
and the target probability:
\begin{align}
\label{eq:interpolation}
 P'(Y_{ij} = y_{ij}|\boldsymbol{x}) = c_{ij}P(Y_{ij} = y_{ij}|\boldsymbol{x}) +
 (1-c_{ij})y_{ij}
\end{align}

When considering segmentation as multi-label classification,
we perform $C$ binary classifications. We do this by using
sigmoid activation function instead of the
softmax activation function at the output
of the network. In this setup, each sample is
simultaneously a positive for its own class and
a negative for the rest of the classes.
This setup also allows us to use outliers during 
training. The ouliers serve as negatives for all
of the $C$ binary classifiers.

To improve the segmentation, we use auxiliary
losses at 4, 8, 16 and 32 times lower resolutions.
The expected output at each resolution is 
a distribution over all segmentation classes. We calculate
this distribution across the corresponding window
from the ground truth labels
at full resolution. We only take into account
pixels that have a valid label. Since we use soft targets
at lower distributions, we use cross entropy
loss between the expected distribution and 
the output of the network at a given resolution.

The model with trainable confidence is trained
by using the confidence loss alongside the multi-class
classifier loss. The confidence loss can be 
interpreted as the negative log-likelihood loss
where the expected output is always 1, that is
to say, the network is always expected to be confident.
\cite{devries18arxiv} show that the confidence tends to 
converge into unity and that confidence training
tends to be a strong regularizer. To prevent the former
they suggest adjusting
$\lambda_{\mathrm{C}}$
so that $\mathcal{L}_{\mathrm{C}}$ tends toward 
a hyperparameter $\beta$ throughout training. 
To minimize the latter, they also suggest applying 
Equation \ref{eq:interpolation} only on
half of the batch at each iteration.
We set $\beta$ to 0.15, and perform
Equation \ref{eq:interpolation} on every second batch.

The model with the separate outlier detection head
is trained with the additional negative log-likelihood loss.
This loss is calculated between the output of 
the second head and the ground truth labels indicating 
whether a pixel is an inlier or an outlier.

The C-way multi-class approach can be trained
to emit uniform distribution at outlier samples.
This is done by minimizing the Kullback-–Leibler (KL)
divergence between the uniform distribution and
the output of the network.

Tables \ref{table:losses} and \ref{table:total-losses}
present an overview of losses used in our experiments.
We use the following assumptions in our notation:
$\boldsymbol{x}$ is an input image, $\boldsymbol{y}$ 
contains ground truth segmentation into $N_{\mathrm{C}}$ classes
and $\boldsymbol{z}$ contains ground truth indicating
whether a pixel is an inlier (label 1) or an outlier (label 0).

Some pixels can be ignored during training
(e.g.the ego-vehicle). To indicate that these pixels should
be ignored, they are given a label greater than the number of
classes $N_{\mathrm{C}}$
in the ground truth segmentation image $\boldsymbol{y}$
and label $2$ in $\boldsymbol{z}$. When training with ouliers,
$\boldsymbol{y}$ can be modified in two different ways:
i) in the C+1-way multi-class setup, $N_{\mathrm{C}}$ is equal to C+1,
the label of the outlier
pixels is set to C+1, while the label of the ignored pixels
needs to be greater than C+1; or ii) the label of the outlier pixel is set 
to any number greater $N_{\mathrm{C}}$.

\clearpage

\begin{table}[htb!]
\centering
\caption{Losses used during training, with the following assumptions:
$\boldsymbol{x}$ is an input image, $\boldsymbol{y}$ contains ground truth segmentation into $N_{\mathrm{C}}$ classes
and $\boldsymbol{z}$ contains ground truth indicating whether a pixel is an inlier or an outlier.
Dimensions of $\boldsymbol{x}$, $\boldsymbol{y}$ and $\boldsymbol{z}$ are $H\times W$.}
\label{table:losses}
\begin{tabular}{p{2cm}|p{10cm}}
  \hline
 Loss & Expression\\
 \hline
  \hline
  \begin{tabular}{p{2cm}}
 multi-class classifier loss\end{tabular} & 
  \begin{tabular}{p{10cm}}
 \begin{gather*}
 \mathcal{L}_\mathrm{MC} = 
 -\mathlarger{\sum}\limits_{i,j \in G_x}
 \lambda_{y_{ij}}
  [\![z_{ij}=1 \wedge y_{ij} \leq N_{\mathrm{C}}]\!]
 \log P(Y_{ij} = y_{ij}|\boldsymbol{x}),\\
 P(Y_{ij} = y_{ij}|\boldsymbol{x}) = \frac{\exp s_{y_{ij}}^{ij}(\boldsymbol{x})}{\sum\limits_{c}\exp{s_{c}^{ij}(\boldsymbol{x})}}
 \end{gather*}
 \end{tabular} 
  \\
   \hline
     \begin{tabular}{p{2cm}}
  multi-label classifier loss \end{tabular}  &
    \begin{tabular}{p{10cm}}
 {\begin{align*}
 \mathcal{L}_\mathrm{ML} = 
 -\mathlarger{\sum}\limits_{i,j \in G_x}
 \mathlarger{\sum}\limits_{c=1}^{N_c}
 [\![y_{ij} \leq N_{\mathrm{C}}]\!]
 \biggl(&[\![y_{ij}\neq c \vee z_{ij}=0]\!]
 \log \frac{1}{1+ \exp s_{y_{ij}}^{ij}(\boldsymbol{x})} \\
 + &[\![y_{ij}=c \wedge z_{ij}=1]\!]
 \log \frac{\exp s_{y_{ij}}^{ij}(\boldsymbol{x})}{1+ \exp s_{y_{ij}}^{ij}(\boldsymbol{x})}\biggr)
 \end{align*}}  \end{tabular} \\
   \hline
   
      \begin{tabular}{p{2cm}}auxiliary loss  \end{tabular}&
      \begin{tabular}{p{10cm}}
      \begin{gather*}
 \mathcal{L}_\mathrm{AUX} =  
 -\mathlarger{\sum}\limits_{r \in R}\mathlarger{\sum}\limits_{i,j \in G_{x}^{r}} [\![N_{ij}^r>\tfrac{r^2}{2}]\!]
 \mathbb{E}_{\boldsymbol{y}_{ij}^r}[\log P(Y_{ij}^r|\boldsymbol{x})],\\
 y_{ijc}^r  = \frac{1}{N_{ij}^r}
 \mathlarger{\sum}\limits_{l=ir}^{ir+r-1}
 \mathlarger{\sum}\limits_{k=jr}^{jr+r-1}
 [\![y_{kl}=c \wedge y_{kl}\leq N_{\mathrm{C}}]\!],\\
 N_{ij}^r =  
 \mathlarger{\sum}\limits_{l=ir}^{ir+r-1}
 \mathlarger{\sum}\limits_{k=jr}^{jr+r-1}
 [\![y_{kl}\leq N_{\mathrm{C}}]\!]
      \end{gather*}
      \end{tabular} 
\\
   \hline
      \begin{tabular}{p{1.5cm}}separate outlier detection head loss  \end{tabular}&
      \begin{tabular}{p{10cm}}
 \begin{align*}
 \mathcal{L}_\mathrm{TH} = 
 -\mathlarger{\sum}\limits_{i,j \in G_x}
 [\![z_{ij} \leq 1]\!]
 \log P(Z_{ij} = z_{ij}|\boldsymbol{x})
 \end{align*} 
 \end{tabular} \\
   \hline
      \begin{tabular}{p{2cm}}Kullback Leibler divergence \end{tabular} &
      \begin{tabular}{p{10cm}}
     \begin{align*}
  \mathcal{L}_\mathrm{KL} = 
  \mathlarger{\sum}\limits_{i,j \in G_{x}}
  [\![z_{ij}=0]\!]\mathrm{KL} (\mathcal{U} \parallel P(Y_{ij}|\boldsymbol{x}))\end{align*}\end{tabular} \\
   \hline
      \begin{tabular}{p{2cm}}confidence loss \end{tabular} &
      \begin{tabular}{p{10cm}}
     \begin{align*}
  \mathcal{L}_\mathrm{C} = 
  -\mathlarger{\sum}\limits_{i,j \in G_{x}}
  [\![y_{ij} \leq N_{\mathrm{C}}]\!]
  \log (c_{ij}|\boldsymbol{x})
  \end{align*}\end{tabular} \\
\hline
\end{tabular}
\end{table}

\clearpage
\begin{table}[htb!]
\centering
\caption{Total training losses}
\label{table:total-losses}
\begin{tabular}{l|c}
  \hline
 Model & Total loss\\
 \hline
  \hline
 C$\times$ multi-class, C+1$\times$ multi-class & 
 $\lambda_\mathrm{MC}\mathcal{L}_\mathrm{MC} + \lambda_\mathrm{AUX}\mathcal{L}_\mathrm{AUX} $ 
 \\
 C$\times$ multi-class with confidence head & 
 $\lambda_\mathrm{MC}\mathcal{L}_\mathrm{MC} +
 \lambda_\mathrm{C}\mathcal{L}_\mathrm{C} + \lambda_\mathrm{AUX}\mathcal{L}_\mathrm{AUX} $
 \\
 C$\times$ multi-label &
 $\lambda_\mathrm{ML}\mathcal{L}_\mathrm{ML} + \lambda_\mathrm{AUX}\mathcal{L}_\mathrm{AUX} $
\\
 C$\times$multi-class with outliers &
 $\lambda_\mathrm{MC}\mathcal{L}_\mathrm{MC} + \lambda_\mathrm{KL}\mathcal{L}_\mathrm{KL} + \lambda_\mathrm{AUX}\mathcal{L}_\mathrm{AUX} $
\\
 two heads &
  $\lambda_\mathrm{MC}\mathcal{L}_\mathrm{MC} + \lambda_\mathrm{TH}\mathcal{L}_\mathrm{TH} + \lambda_\mathrm{AUX}\mathcal{L}_\mathrm{AUX} $
\\
\hline
\end{tabular}
\end{table}
\subsection{Validation of Dense Oultier Detection Approaches}
Table \ref{table:OOD_detection} is an extension of Table 2
from the main paper. It provides results of outlier detection
and semantic segmentation for a model with a separate
confidence head \cite{devries18arxiv} and a semantic segmentation
model trained using Monte Carlo (MC) dropout after each of the dense
and upsampling layers, with epistemic uncertainty
used as a criterion for outlier detection \cite{kendall17nips,kendall15arxiv}.
The MC-dropout is also used during the inference phase,
meaning that the final output of the model is the mean value of
50 forward passes.
Both of the models trained with
dropout perform outlier detection
better than the baseline model. They, however,
perform worse than any of the
models trained with noisy negatives.
Out of all of the tested models, 
the model with the confidence head
provides the worst
outlier detection accuracy.
We observe a drop in semantic
segmentation accuracy on WildDash 
validation dataset
when compared to the baseline model
for all of the additional models.

\begin{table}[htb!]
\centering
\caption{Validation of dense 
  outlier detection approaches.
  WD denotes WildDash val.}
\label{table:OOD_detection}
\begin{tabular}{|c|c||c|c|c|}
  \hline
  Model & ImageNet &
    \multicolumn{1}{c|}{AP WD-LSUN} &
    \multicolumn{1}{c|}{AP WD-Pascal} &
    \multicolumn{1}{c|}{mIoU WD}\\
 \hline

 \hline


  \hline
  \multicolumn{1}{|l|}{
    C$\times$ multi-class, dropout 0.2} & 
    \ding{55} & $64.09 \pm 0.97 $ & 11.87 & 49.13 
    \\
      \hline
  \multicolumn{1}{|l|}{
    C$\times$ multi-class, dropout 0.5} & 
    \ding{55} & $ 63.37 \pm 1.13$ & 13.18 & 47.25
    \\
   \hline
  \multicolumn{1}{|l|}{
    confidence head} & 
    \ding{55} & $54.40 \pm 0.80$ & 4.41 & 45.38
    \\
\hline
\end{tabular}
\end{table}

\begin{figure}[htb!]
  \centering
  \includegraphics[width=0.24\textwidth]{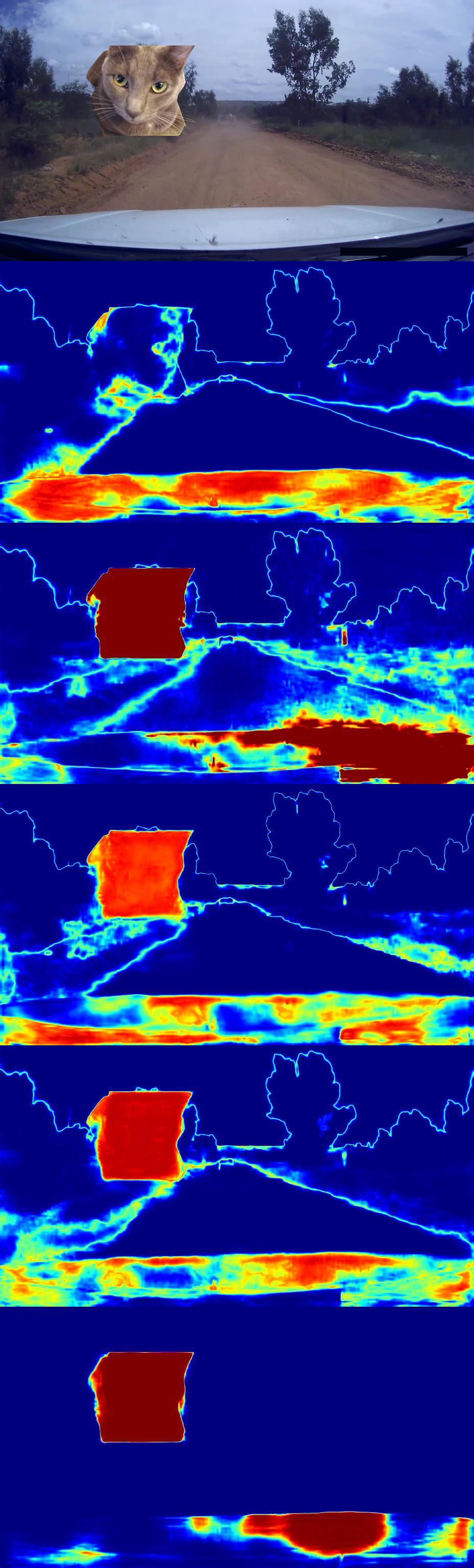}
  \includegraphics[width=0.24\textwidth]{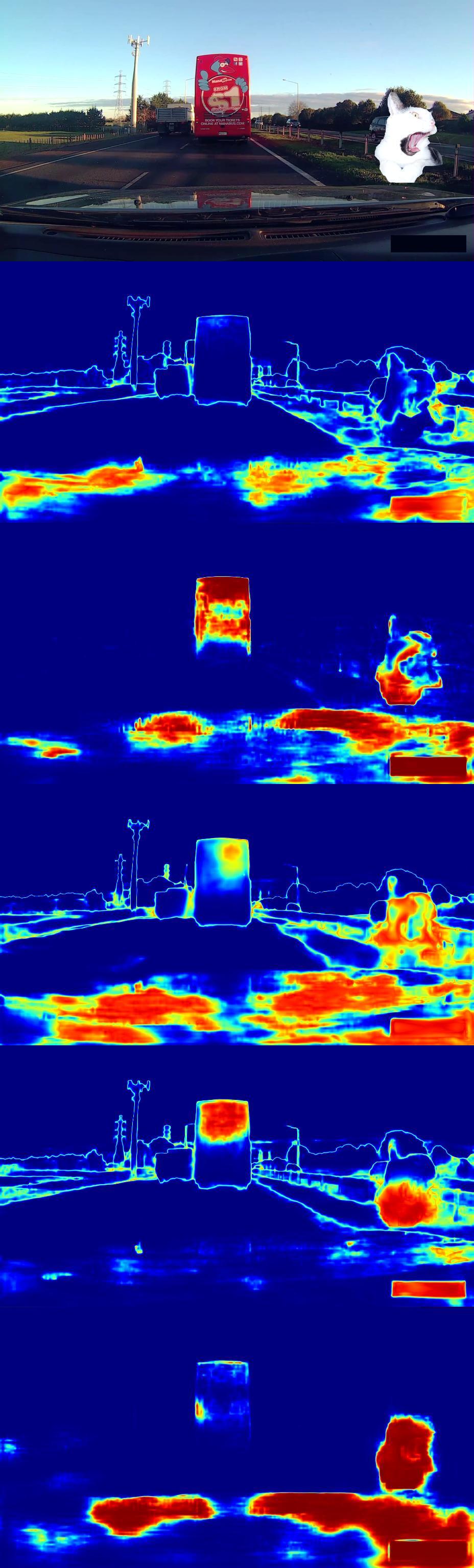}
  \includegraphics[width=0.24\textwidth]{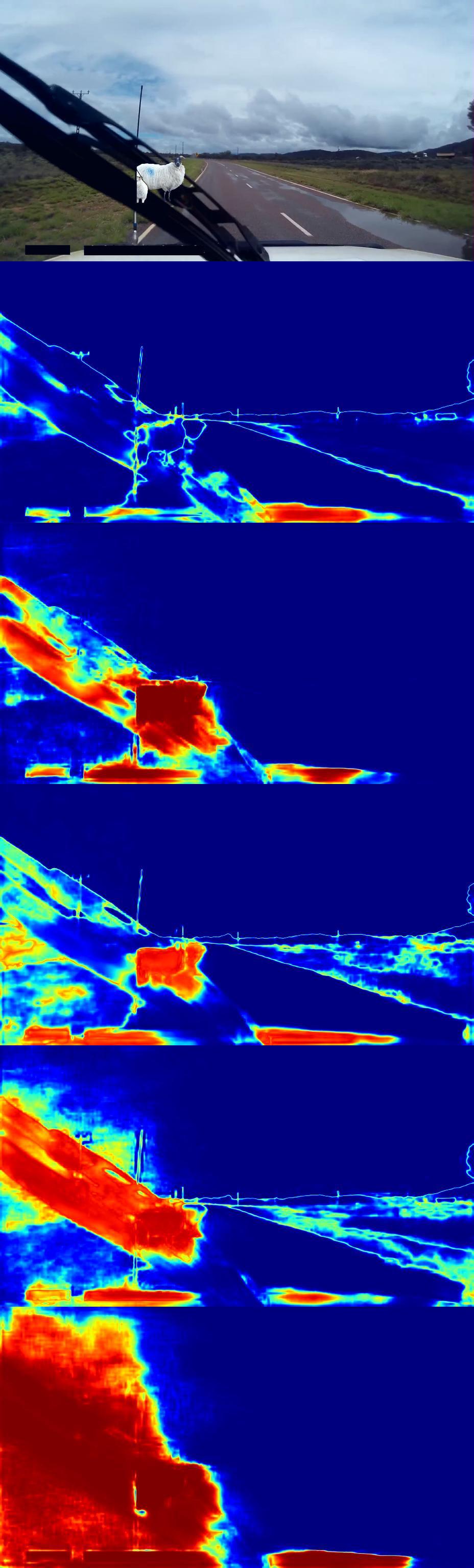}
  \includegraphics[width=0.24\textwidth]{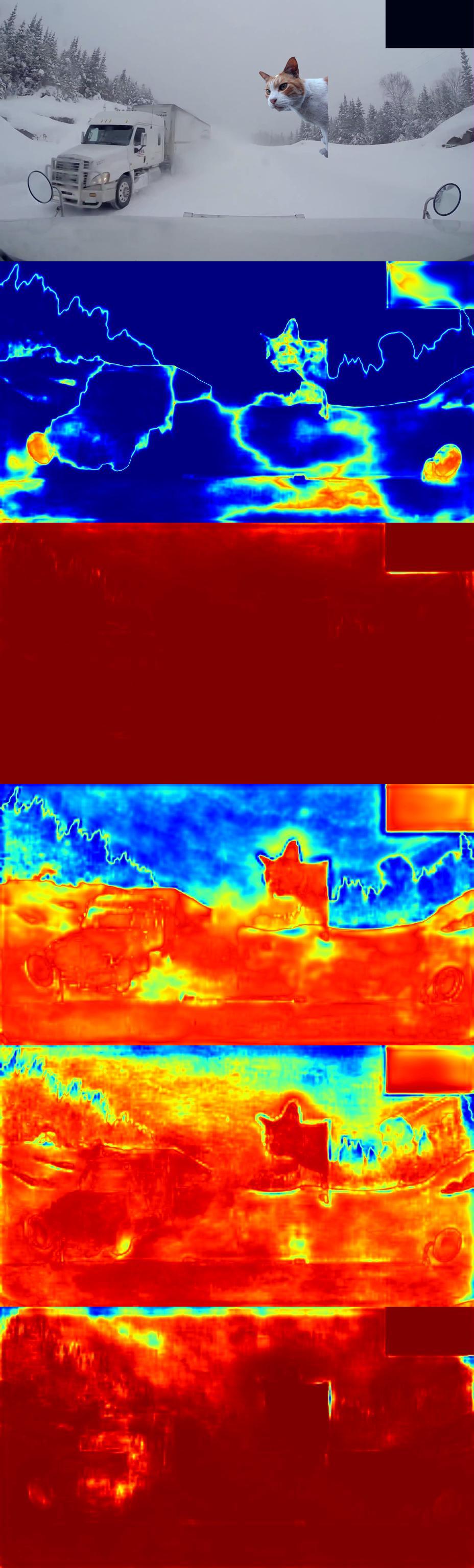}
  \caption{Dense outlier detection 
  with models presented in Table 2.
  Row 1 shows Wilddash val images with 
  pasted PASCAL VOC 2007 animals. 
  Row 2 shows outlier probabilities 
  obtained with the C-way multi-class model
  trained without negatives.
  Red colour indicates a high probability 
  that a pixel is an outlier.
  Subsequent rows correspond 
  to models trained with noisy negatives:
  the C+1-way multi-class model (row 3), 
  the C-way multi-label model (row 4), 
  the C-way multi-class model (row 5) and 
  the two-head model (row 6). 
  }
  \label{fig:pw_ood_det_models}
\end{figure}

Figure \ref{fig:pw_ood_det_models} gives a qualitative
insight into the results from Table 2 of the main paper.
It explores validation performance of different
outlier detector variants described in Section 3.2
of the main paper.

Row 1 contains images from WD-Pascal.
Each subsequent row shows the output of a
different outlier detector.
Row 2 corresponds to
the C-way multi-class model
trained without negative samples (baseline). 
This model assigns high
outlier probabilty at semantic borders. 
Consequently it detects borders of outlier patches. 
It is however unable to detect entire outlier objects.
Since the baseline model with ODIN,
the model with the confidence head
and the models trained with dropout
perform qualitatively similarly,
we omit those results for brevity.

Row 3 shows the output of the C+1-way multi-class model.
This model had the lowest AP score. 
Qualitative analysis suggests that this model 
makes more false outlier predictions
which in turn lowers the AP score.

Row 4 shows the C-way multi-label model and row 5
shows the C-way multi-class model. They perform
similarly, though the C-way multi-label model seems
to be more robust to domain shift (cf. column 4).
These models are more successful at outlier detection 
than the baseline model, though they also 
assign a high outlier probability to border pixels.

Row 6 shows the output of the two-head model. 
It is successful at detecting 
outliers without falsely detecting 
borders as outliers. Furthermore, its detections 
tend to be coarser than in other models.

All models trained with ImageNet-1k-bb have a problem
with large domain shift as exemplified in column 4.
This indicates that it would be very hard 
to achieve high AP scores on WD-Pascal
since images with large domain shift are perceived
as equally foreign as the pasted Pascal objects.
Column 3 shows that models trained using negative data
classify the windshield wiper as outlier. This is 
interesting when compared to the model which has 
seen WildDash validation during training (row 3 of 
Figure \ref{fig:bench_bin_oe_outlier_inlier}). This
is further discussed in Section 3.

\subsection{Influence of the Training Data}
\begin{figure}[htb]
  \centering
  \includegraphics[width=0.98\textwidth]{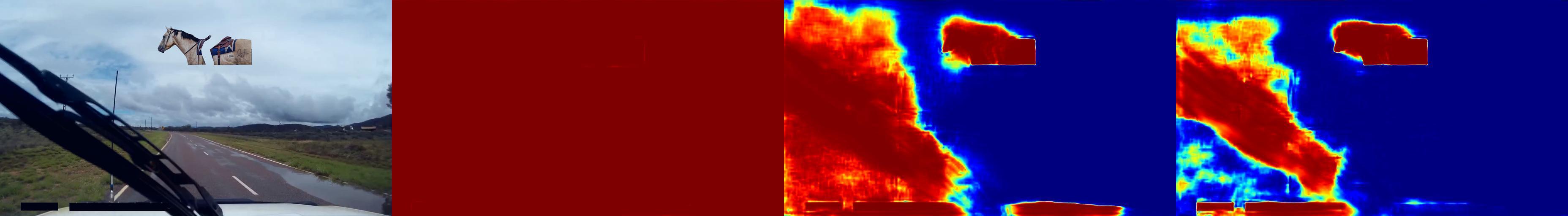}\\
  \includegraphics[width=0.98\textwidth]{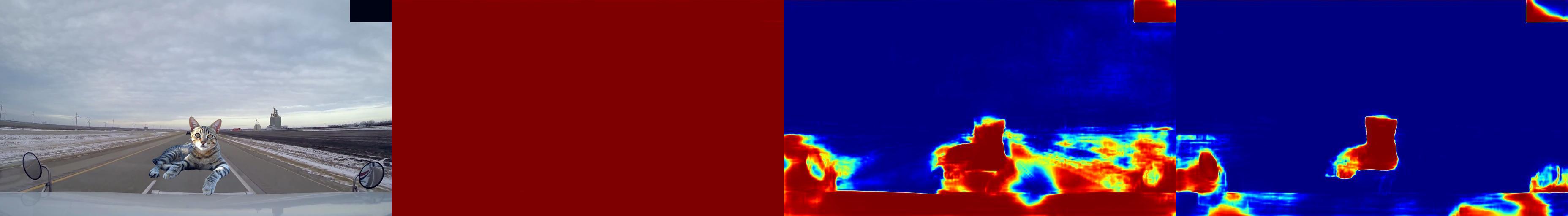}\\
  \includegraphics[width=0.98\textwidth]{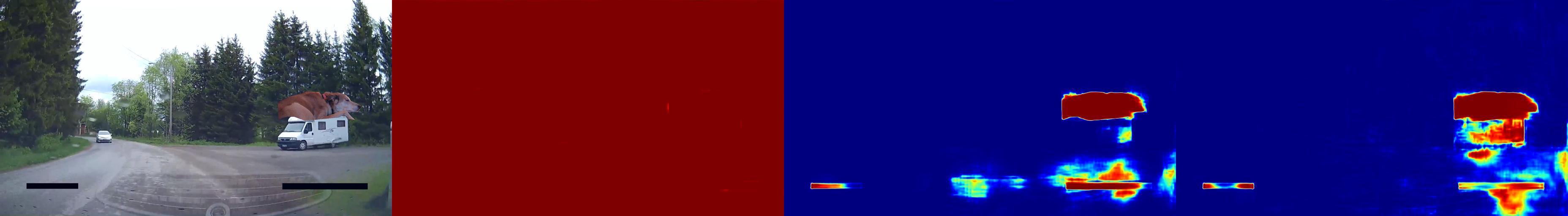}\\
  \includegraphics[width=0.98\textwidth]{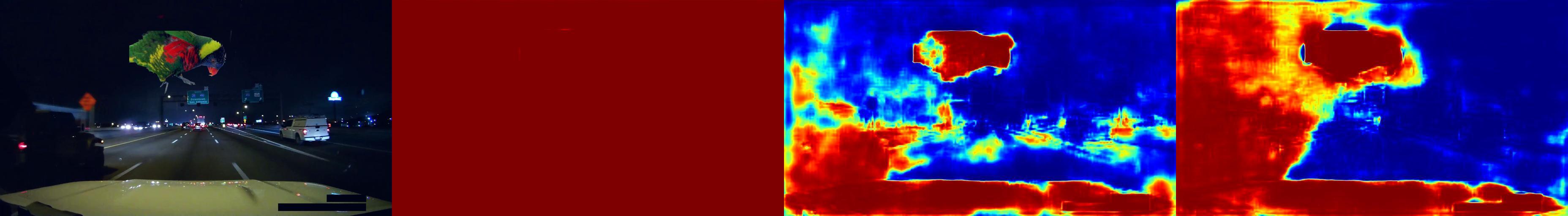}
  \caption{Outlier detection with two-head models
    trained on different inlier datasets.
    All models have been trained 
    with pasted noisy negatives 
    from ImageNet-1k-bb
    as presented in from Table 4.
    Column 1 contains Wilddash images 
    with pasted PASCAL VOC 2007 animals.
    Columns 2-4 show predictions of models
    trained on Cityscapes, Vistas, 
    and Cityscapes and Vistas, respectively.
    Red colour indicates a high probability 
    that a pixel is an outlier.
  }
  \label{fig:pw_ood_det_in_data}
\end{figure}

Figure \ref{fig:pw_ood_det_in_data} illustrates the
validation performance of the two-head model
depending on the inlier training dataset, which 
was shown in Table 4 of the paper.
Column 1 shows four validation images.

Column 2 presents the corresponding results 
of the two-head model trained on Cityscapes.
This model classifies
all of the WildDash pixels as outliers. 
This indicates that models trained on Cityscapes are
very sensitive to domain shift. 

Column 3 shows that the two-head model trained
on Vistas dataset is significantly better
at outlier detection. This improvement indicates that 
models trained on Vistas show more resilience
to domain shift.

Column 4 depicts a model trained 
on both Vistas and Cityscapes. 
It performs similarly to the model 
trained on Vistas.

Interestingly, the model
trained on Vistas shows more resilience to unusual
conditions (dark image in row 4, unusual vehicle in 
row 3), while the model trained on combined datasets 
shows more precision in normal situations
(grass in row 1, road in row 2).

\begin{figure}[htb]
  \centering
  \includegraphics[width=0.98\textwidth]{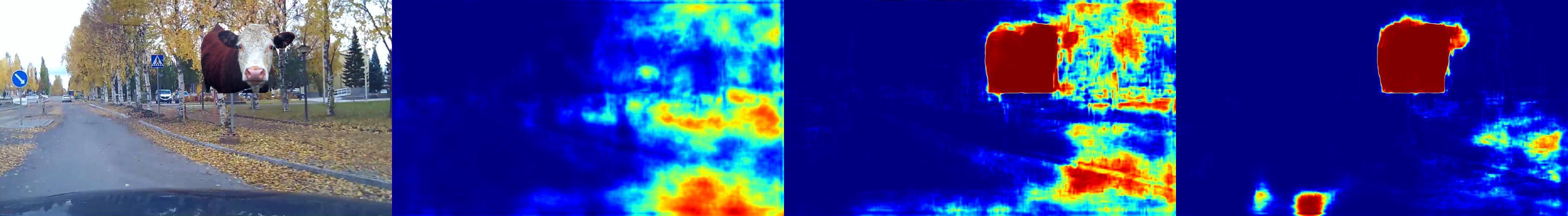}\\
  \includegraphics[width=0.98\textwidth]{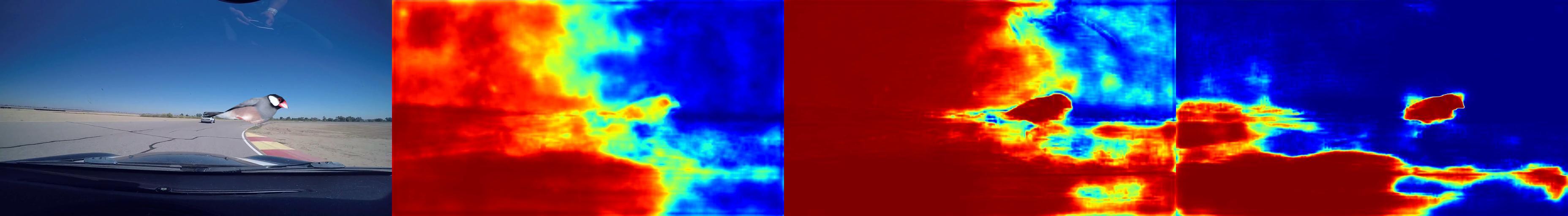}\\
  \includegraphics[width=0.98\textwidth]{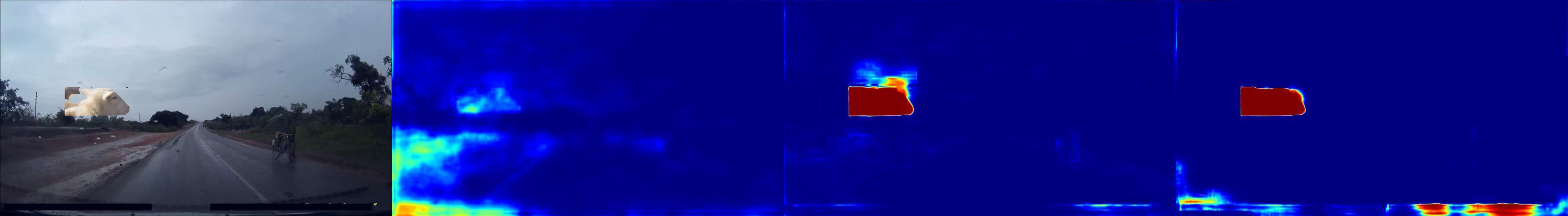}\\
  \includegraphics[width=0.98\textwidth]{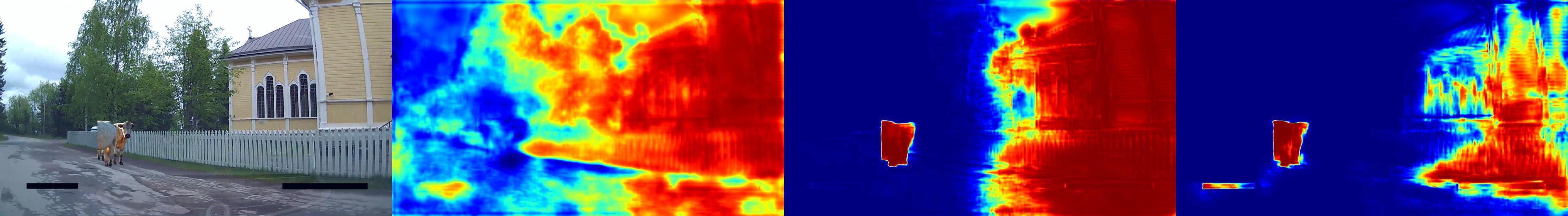}
  \caption{Outlier detection with two-head models
    trained on different negative datasets.
    All models have been trained 
    by pasting negatives into 
    inliers from Vistas
    as presented in Table 5.
    Column 1 shows original Wilddash images 
    with pasted PASCAL VOC 2007 animals.
    Columns 2-4 show predictions of models
    trained with noisy negatives
    from Imagnet-1k-full (without pasting), 
    ImageNet-1k-full (with pasting), 
    and ImageNet-1k-bb (with pasting),
    respectively.
    Red colour indicates a high probability 
    that a pixel is an outlier.
  }
  \label{fig:pw_ood_det_out_data}
\end{figure}

Figure \ref{fig:pw_ood_det_out_data} shows the
results of the two-head model depending on 
the negative training dataset (cf. Table 5).

The first column shows the validation images.

The second column shows the model trained 
using ImageNet-1k-full without pasting.
It tends to make coarse 
predictions. Consequently it is unable to
detect small outlier
patches.

The third column illustrates the model trained
using ImageNet-1k-full with pasting. It is better
at detecting outlier samples. It is however
sensitive to domain shift. This is because
of the increased overlap between positive and
negative images (in classes such as sky,
vegetation or road which appear often
in the backgrounds of ImageNet images).

The last column shows the model trained using
ImageNet-1k-bb with pasting which performs
the best both qualitatively and quantitatively.

\subsection{Comparing the Two-Head and 
  C-way Multi-class Models}
Figure \ref{fig:bench_bin_oe_normal} accompanies Table 1
of the main paper. It presents the combined output
of the two-head model (column 2) and the C-way model trained with
ImageNet-1k images (column 3) on images from 
WildDash test set (column 1).

The first four rows show images taken in normal
conditions, while the last two rows
show oulier images.

The C-way model tends to classify
small objects (cf. poles in image in row 3)
as well as distant objects (cf.
trucks in the distance in the image in the third row) 
as outliers. This model makes
more false outlier detections
in typical traffic scenes. 

Furthermore, the two-head model performs 
better in negative images.
Output of the C-way model on negative images
contains small patches not classified as outliers.

\newcommand{\mywt}{0.85\textwidth}
\begin{figure}[htb]
  \centering
  \includegraphics[width=\mywt]{wd0001final.jpg}
  \includegraphics[width=\mywt]{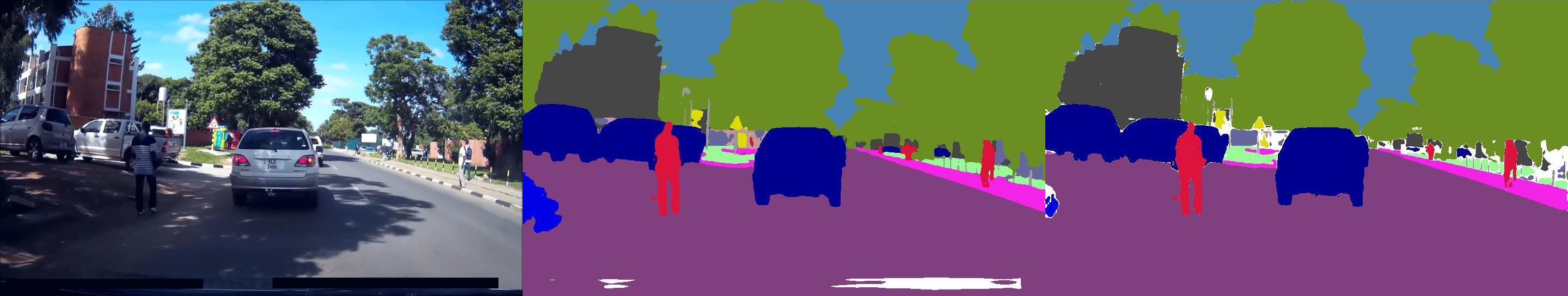}
  \includegraphics[width=\mywt]{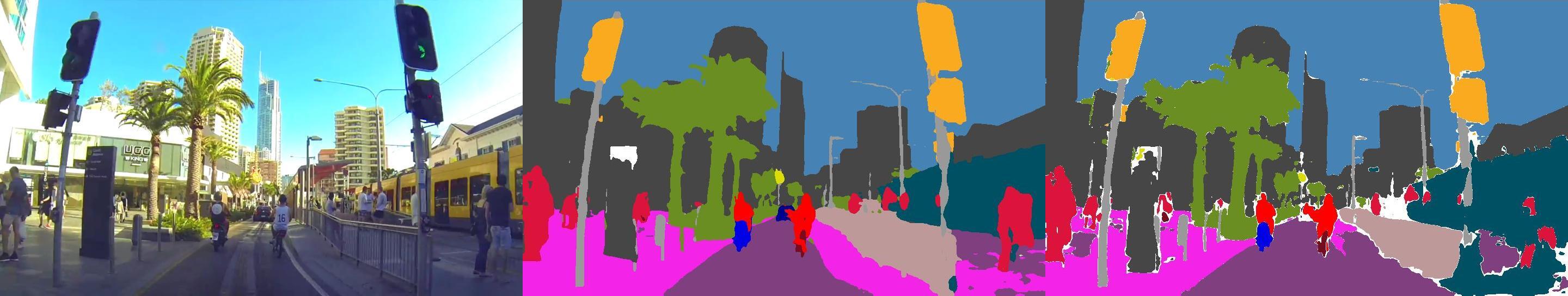}
  \includegraphics[width=\mywt]{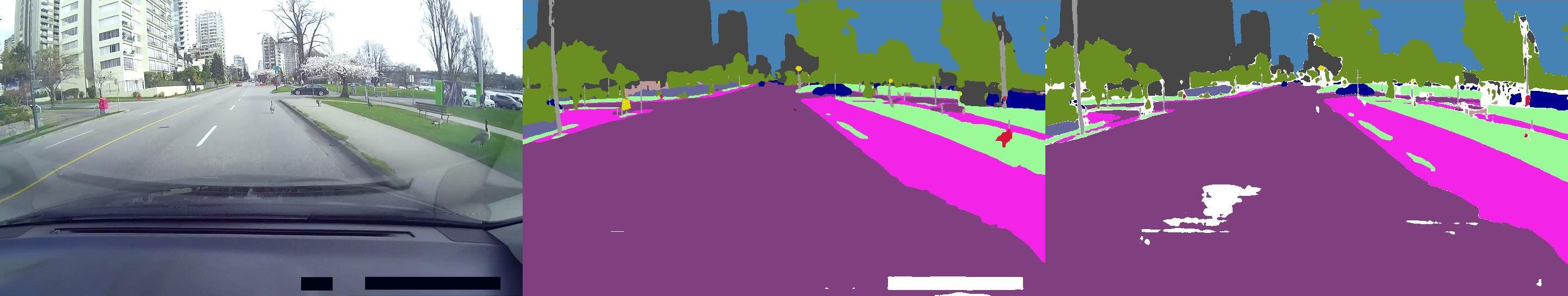}
  \includegraphics[width=\mywt]{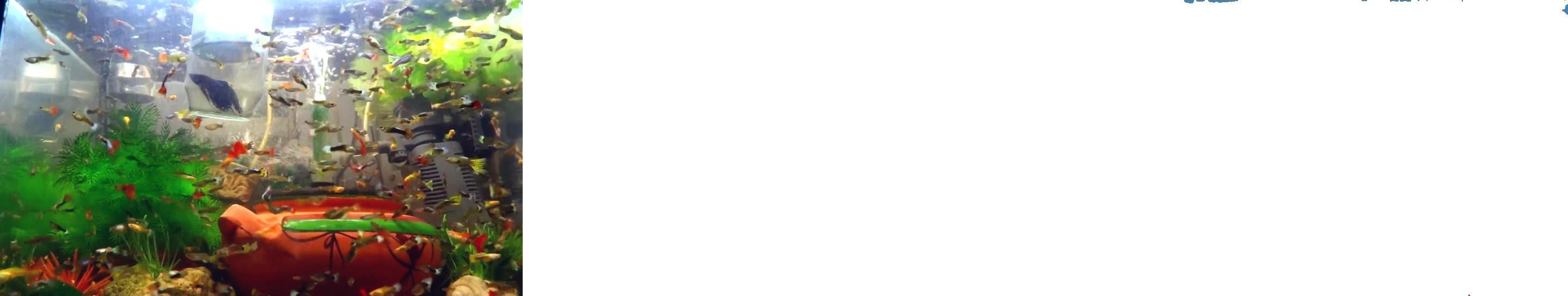}
  \includegraphics[width=\mywt]{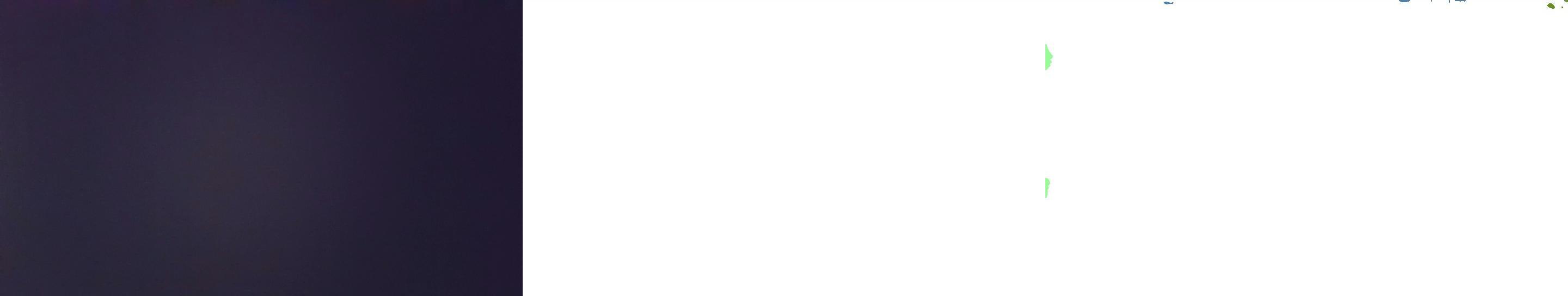}
  \caption{Qualitative performance 
    on the WildDash benchmark.
    Each triplet contains 
    a test image (left),
    the output of the two-head model (center), 
    and the output of the C-way multi-class model 
    trained to predict uniform distribution 
    in outliers (right).
    The two-head model does not produce
    false positives at semantic borders
    (rows 1-4).
    Both models correctly recognize
    outliers in negative images (rows 5-6).
  }
  \label{fig:bench_bin_oe_normal}
\end{figure}

\begin{figure}[htb]
  \centering
  \includegraphics[width=\mywt]{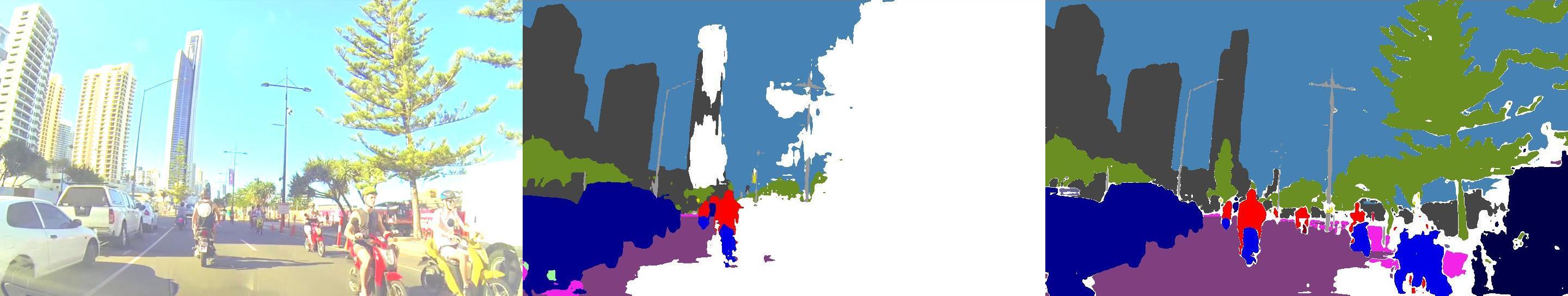}
  \includegraphics[width=\mywt]{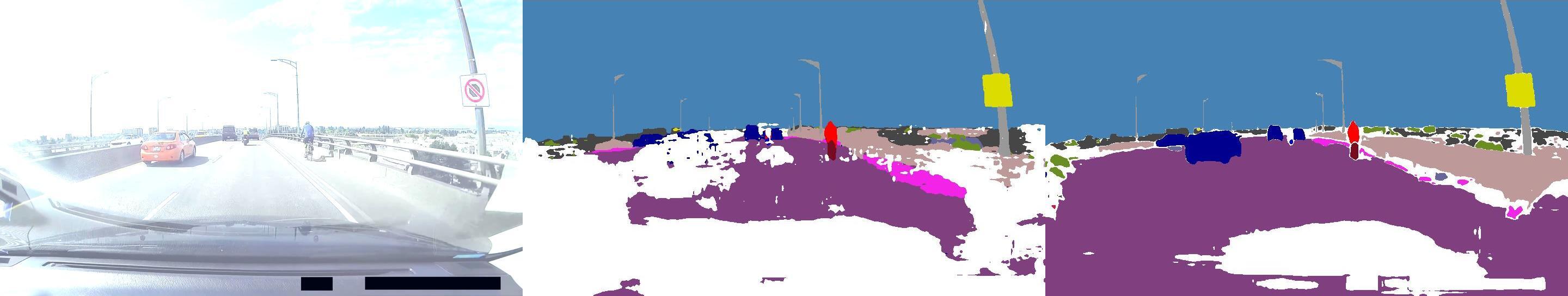}
  \includegraphics[width=\mywt]{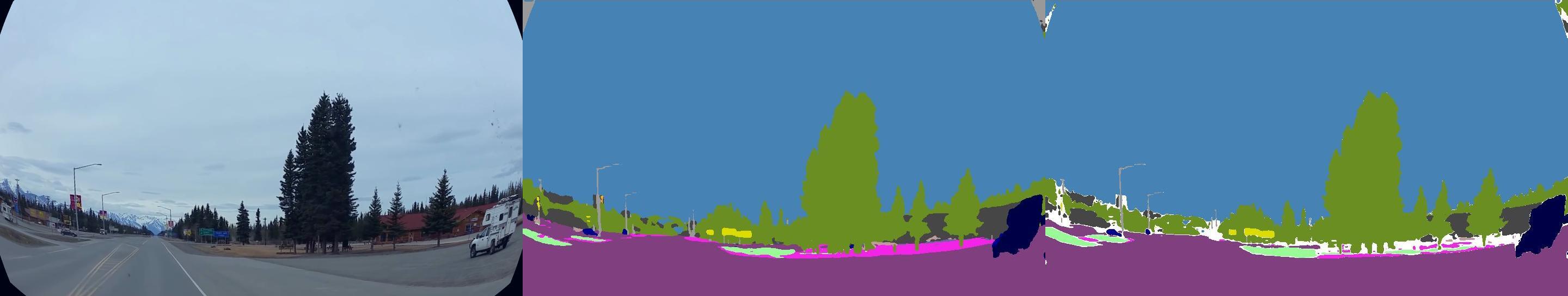}
  \includegraphics[width=\mywt]{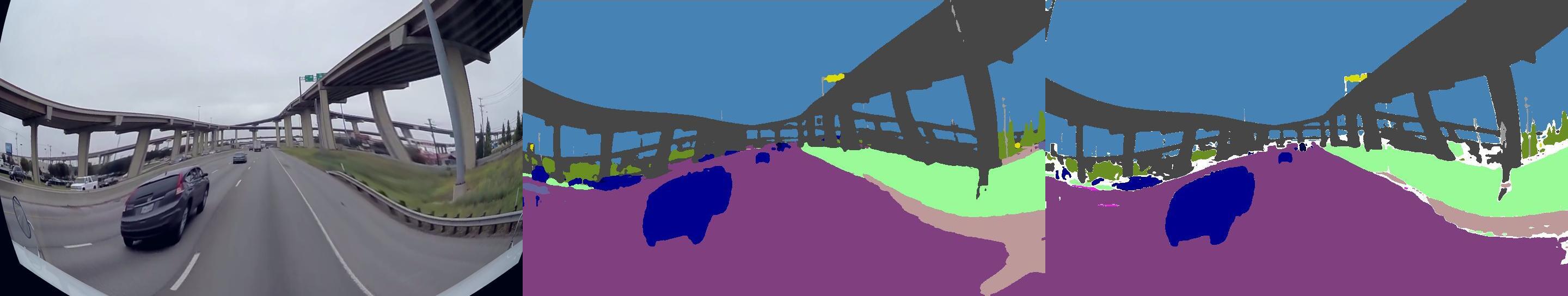}
  \includegraphics[width=\mywt]{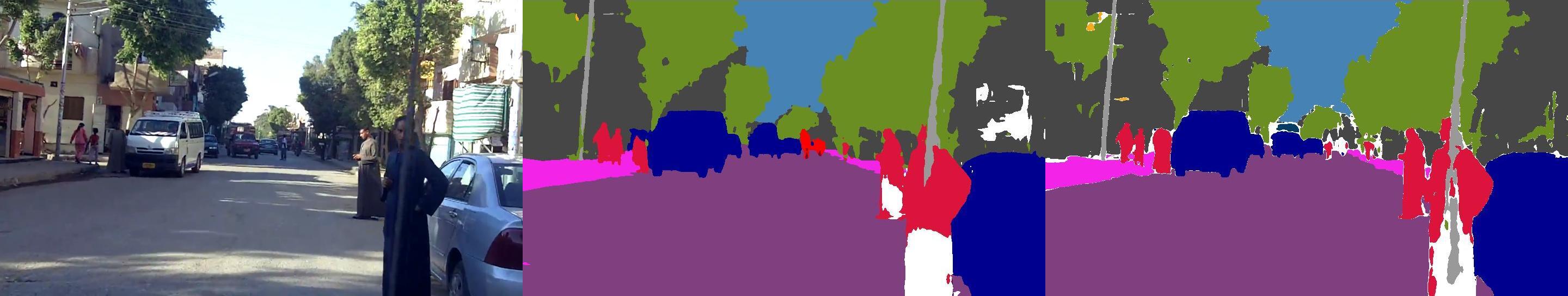}
  \includegraphics[width=\mywt]{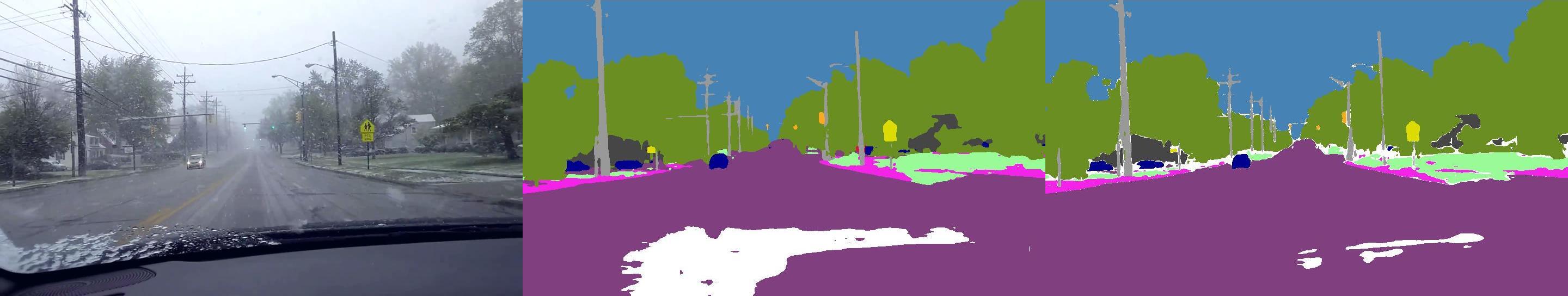}
  \includegraphics[width=\mywt]{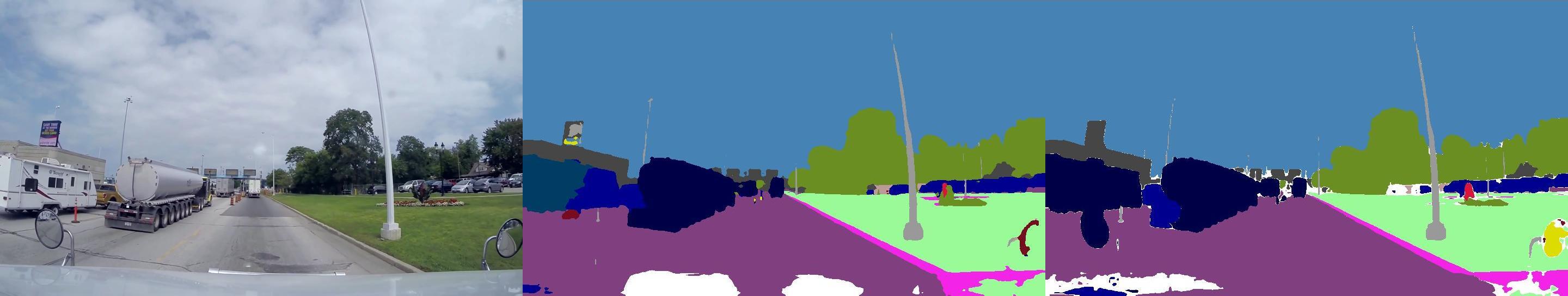}
  \caption{Qualitative performance 
    on WildDash test images 
    with overexposure 
    and distortion hazards.
    Each triplet contains 
    a test image (left),
    the output of the two-head model (center), 
    and the output of the C-way multi-class model 
    trained to predict uniform distribution 
    in outliers (right).
    The two-head model rejects
    images with severe hazards.
  }

  \label{fig:bench_bin_oe_hazard}
\end{figure}

Figure \ref{fig:bench_bin_oe_hazard} further
clarifies Table 6 from the main paper.
It illustrates impact of WildDash hazards 
on the output of the submitted models.

The results show that the two-head model
is more sensitive to overexposure and distortion
hazards, though it succeeds when the hazard
is not severe (rows 1-4).

Row 5 contains an example of occlusion. Both
of the models are able to successfully 
segment the torso or the person behind the pole but 
they struggle with the lower part.
Row 6 contains an example of the particles hazard
(rain on the windshiled), while
row 7 contains an example of the variation hazard
(with tank truck being an atypical example of a truck).

\begin{figure}[htb]
  \centering
  \includegraphics[width=\mywt]{wd0013final.jpg}
  \includegraphics[width=\mywt]{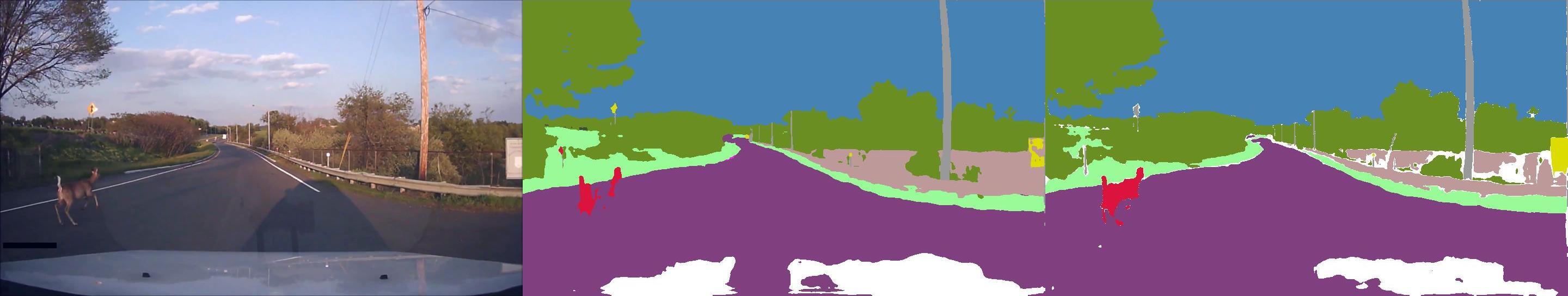}
  \includegraphics[width=\mywt]{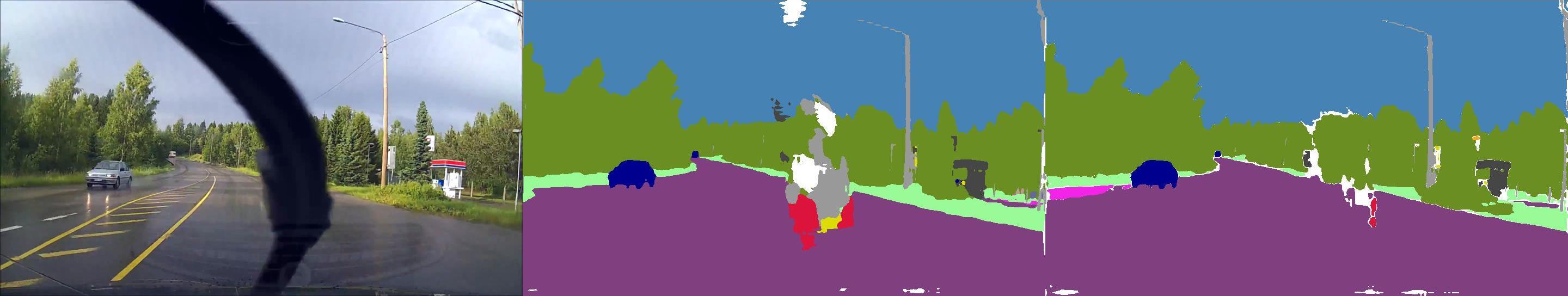}
  \caption{Qualitative performance 
  of our two submissions 
  to the WildDash benchmark.
  Each triplet contains 
  a test image (left),
  the output of the two-head model (center), 
  and the output of the model trained 
  to predict uniform distribution in outliers (right).
  Rows 1-2 show that our current models
  are unable to correctly detect small outlier objects. 
  Row 3 shows that the windshield wiper
  is recognized as inlier, which occurs
  due to its presence in WildDash val
  (cf.\ Figure \ref{fig:pw_ood_det_models}).
  }
  \label{fig:bench_bin_oe_outlier_inlier}
\end{figure}

Figure \ref{fig:bench_bin_oe_outlier_inlier} expands
on the failure cases shown in Figure 4 of the main paper
which are worth exploring in future work.

The first two rows show images with animals
on roads. Animals are usually classified
as pedestrians. The two-head model classifies most
of the pixels of the image in row 1 as outliers.
Both of the models fail to classify the 
animals as outliers accurately.

The windshield wiper in the row 3 
is classified as an inlier. 
This is because the submitted models were trained 
on Wilddash val which contains examples of images 
with windshields wipers. 
Those pixels are ignored during training but 
they still influence the features 
extracted by the dense feature extractor.
This view is supported by
Figure \ref{fig:pw_ood_det_models}
where images in column 3 demonstrate
that a model trained only on Vistas 
classifies windshield wipers as outliers.
Handling ignored training pixels is a 
suitable direction for future work.

\end{document}